\documentclass{article}

\usepackage{arxiv}

\usepackage[utf8]{inputenc} 
\usepackage[T1]{fontenc}    
\usepackage{hyperref}       
\usepackage{url}            
\usepackage{booktabs}       
\usepackage{amsfonts}       
\usepackage{nicefrac}       
\usepackage{microtype}      
\usepackage{amsmath}
\usepackage{cleveref}       
\usepackage{lipsum}         
\usepackage{graphicx}
\usepackage{doi}

\title{Attacks on multimodal models}

\newif\ifuniqueAffiliation
\uniqueAffiliationtrue

\ifuniqueAffiliation 
\author{\hspace{1mm}Iablochnikov Viacheslav
        \\
	Department of Physics\\
	National Research University Higher School of Economics\\
	Moscow, Russia \\
	\texttt{vsyablochnikiv@edu.hse.ru} \\
	\And
	\hspace{1mm}Alexander Rogachev \\
	Department of Computer Science\\
	National Research University Higher School of Economics\\
	Moscow, Russia \\
	\texttt{airogachev@hse.ru} \\
}
\else
\usepackage{authblk}

\newbox{\orcid}\sbox{\orcid}{\includegraphics[scale=0.06]{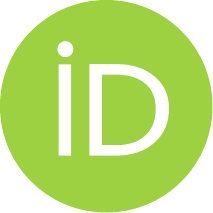}} 
\author[1]{%
	\usebox{\orcid}\hspace{1mm}Iablochnikov Viacheslav\thanks{\texttt{vsyablochnikiv@edu.hse.ru}}%
}
\author[1,2]{%
	\usebox{\orcid}\hspace{1mm}Alexander Rogachev\thanks{\texttt{airogachev@hse.ru}}%
}
\affil[1]{National Research University Higher School of Economics, Moscow, Russia}
\affil[2]{National Research University Higher School of Economics, Moscow, Russia}
\fi

\hypersetup{
pdftitle={Attacks on multimodal models},
pdfsubject={q-bio.NC, q-bio.QM},
pdfauthor={Iablochnikov Viacheslav, Alexander Rogachev},
pdfkeywords={Vision Language model (pretrained VLM), CLIP, input-based adversarial attacks, Universal Adversarial perturbations (UAPs), patch attacks},
}

\usepackage{url}

\usepackage{amssymb}
\usepackage{amsthm}
\usepackage{mathrsfs}
\usepackage{multicol}
\usepackage{svg}
\usepackage{xcolor}
\usepackage{multirow}
\usepackage{colortbl}
\usepackage{tabularx}
\usepackage{pifont} 
\usepackage{nameref}
\usepackage{hyperref}

\usepackage{tcolorbox}
\definecolor{bg}{gray}{0.95}

\begin{document}
\maketitle

\begin{abstract}
Today, models capable of working with various modalities simultaneously in a chat format are gaining increasing popularity. Despite this, there is an issue of potential attacks on these models, especially considering that many of them include open-source components. It is important to study whether the vulnerabilities of these components are inherited and how dangerous this can be when using such models in the industry. This work is dedicated to researching various types of attacks on such models and evaluating their generalization capabilities. Modern VLM models (LLaVA, BLIP, etc.) often use pre-trained parts from other models, so the main part of this research focuses on them, specifically on the CLIP architecture and its image encoder (CLIP-ViT) and various patch attack variations for it. Code is available here: \href{https://github.com/slava-qw/image-retrieval-robustness}{https://github.com/slava-qw/image-retrieval-robustness}
\end{abstract}

\section*{Introduction}

The rising popularity of large language models (LLMs) (\cite{LLM}) has fueled rapid growth in multimodal "image2text" models, capable of processing both images and text simultaneously (\cite{VLM}). Today, these models are not only attracting the attention of researchers, but are also widely used across various industries. The integration of text and image processing within a single model offers significant opportunities for applying such approaches to a wide range of tasks. However, this also opens up new avenues for attacks on these models, as many of them consist of pretrained components or rely on open-source solutions, thereby inheriting associated vulnerabilities.

Models used in practical applications must be robust against various input-based attacks. Specifically, compromised input data should not have the ability to manipulate the model's behavior in undesirable ways—such as leaking personal information, altering the behavior of public services, or assisting users in committing illegal activities.

One of the most popular forms of input attacks involves adding specific noise to the original image (\cite{Good_14}, \cite{Madry_18}, \cite{FGSM_fut}). While these pixel-level attacks are powerful (\cite{pgd_expl}), they lack generalizability, as a unique perturbation needs to be crafted for each individual image (\cite{Madry_18}). As a result, methods capable of generalizing across multiple images have gained prominence. Among these methods, patch attacks and universal adversarial perturbations (UAPs) hold a special place (\cite{patch_survey}, \cite{UAPs_survey}).

\paragraph{Universal Adversarial Perturbations (UAPs)} are a type of attack where a single perturbation is created to mislead the model across a wide range of different images \cite{UAPs}. UAPs are effective due to their simplicity and versatility, as the same perturbation can be applied to multiple images, leading to incorrect predictions by the model. These attacks are particularly dangerous for multimodal models \cite{VLM_UAPs}, which leverage a combination of various data types, since they allow an attacker to encode specific harmful information via these perturbations.

\paragraph{Patch attacks} are a method derived from UAPs, in which a specific patch (a region of the image) is altered to mislead the model \cite{patch}. This patch can be applied to different images and will be equally effective regardless of the context in which it is used \cite{hijacks}. This makes patch attacks a powerful tool, as they can target multimodal models handling various types of data, such as images, text, or audio.

Both techniques pose a significant threat to the reliability of multimodal models, as they allow adversaries to bypass security mechanisms and force models to make incorrect decisions. Understanding and developing methods to defend against such attacks is crucial, as it will enhance the robustness of deep neural networks and ensure their safe application in critical areas.

\section*{Related works}
\subsection*{Multimodal models}

The rapid growth in popularity and various methods of language processing have made it possible to combine two modalities: text and images, to create multimodal models (VLMs). These models primarily build on the success of large language models \cite{lmsys_arena} (for example, \cite{llama}), expanding their capabilities to the visual domain. To reduce computational costs, modern models use pre-trained components. Their results are combined using a trainable neural network, which can be represented either by additional layers \cite{open_flamingo}, or a learnable mapping between the two modalities \cite{BLIP}.

\subsection*{CLIP}

One type of Vision-Language Model (VLM) is the Contrastive Language-Image Pre-training (CLIP) model developed by OpenAI \cite{CLIP}. CLIP employs contrastive learning, which is based on the principle of bringing similar elements closer together and pushing dissimilar ones apart in the representation space. It uses InfoNCE as its loss function, a variation of Noise Contrastive Estimation (NCE), which applies a softmax function and a temperature parameter to compute similarities between positive pairs (an image and its caption) and negative pairs (the same image with captions from other images in the batch).

$$
\mathcal{L_{\text{clip}}} = -\frac{1}{2|\mathcal{B}|} \sum_{i=1}^{|\mathcal{B}|} \Bigg(\overbrace{\log \frac{e^{t \mathbf{x}_i \cdot \mathbf{y}_i}}{\sum_{j=1}^{|\mathcal{B}|} e^{t \mathbf{x}_i \cdot \mathbf{y}_j}}}^{\text {image } \rightarrow \text { text softmax }} + \overbrace{\log \frac{e^{t \mathbf{x}_i \cdot \mathbf{y}_i}}{\sum_{j=1}^{|\mathcal{B}|} e^{t \mathbf{x}_j \cdot \mathbf{y}_i}}}^{\text {text } \rightarrow \text { image softmax }}\Bigg)
$$

where $\mathbf{x}_i=\frac{E_{\text{img}}\left(I_i\right)}{\left\|E_{\text{img}}\left(I_i\right)\right\|_2}, \
\mathbf{y}_i=\frac{E_{\text{txt}}\left(T_i\right)}{\left\|E_{\text{txt}}\left(T_i\right)\right\|_2}$, $|\mathcal{B}|$ - batch size.

The architecture of CLIP includes two encoders—one for images and one for text—that are trained to align their inputs in a shared embedding space. Trained on a dataset of 400 million image-caption pairs, CLIP demonstrates zero-shot capabilities for classifying images and generating corresponding text. For example, CLIP based on ResNet-101 \cite{ResNet_101} achieves an accuracy of $76.2\%$ in image classification tasks.

It’s worth noting that researchers continue to improve upon the CLIP approach. For instance, SigLIP \cite{Siglip} adapts CLIP’s multi-class problem to a binary cross-entropy loss function, enhancing zero-shot performance even with smaller batch sizes.

$$
\mathcal{L_{\text{SigLIP}}} = -\frac{1}{|\mathcal{B}|} \sum_{i=1}^{|\mathcal{B}|} \sum_{j=1}^{|\mathcal{B}|} \underbrace{\log \frac{1}{1+e^{z_{i j}\left(-t \mathbf{x}_i \cdot \mathbf{y}_j+b\right)}}}_{\mathcal{L}_{i j}}
$$
here, \(z_{ij}\) is the label for a given image and input text pair, equal to 1 if the pair is positive, and -1 otherwise. The parameters \(t\) and \(b\) are trainable.

Another variation, \cite{Llip}, introduces a cross-attention module to account for image encoding based on target captions, further improving benchmark results.

\subsection*{Adversarial attacks}
Let \( x_0 \in \mathbb{R}^d \) be the original data. An attack is considered successful if, after applying a predefined transformation \( \mathcal{A} \), the model \( f_{\theta} \) produces the desired result on the attacked images \( x = \mathcal{A}(x_0) \). 

For classification tasks, if \( C_i \) and \( C_t \) are the original and target labels, then \( f_{\theta}(x) = C_t \neq C_i \). The transformation \( \mathcal{A} \) is typically chosen to be an additive function in the pixel space for simplicity (although other methods exist \cite{NCF}). To make the attack less noticeable, a constraint is imposed on the transformation, ensuring that \( \|x - x_0\|_p \leq \varepsilon \).

This optimization problem is often analytically intractable, so gradient-based methods are employed to solve it. Similar to gradient descent, the gradient of the loss function with respect to the added noise is computed and its values are updated iteratively. This method is known as the Fast Gradient Sign Method (FGSM) \cite{Good_14}. However, one step of FGSM is often insufficient to achieve the desired result, so this process can be repeated several times:
$$
x_{k+1} = \mathcal{P}_S (x_k + \alpha \cdot \text{sign} (\nabla_x L (\theta, x_k , y)))
$$
where \( \mathcal{P}_S \) is a projection operator and \( L \) is the loss function. This iterative method is known as Projected Gradient Descent (PGD) \cite{Madry_18}. These methods and their subsequent variations \cite{FGSM_fut} are the most popular solutions for optimization tasks in the context of adversarial attacks, as they are computationally efficient.

For UAPs (Universal Adversarial Perturbations), the methods for generating adversarial images and solving the optimization problem remain nearly the same, but an additional constraint is introduced: the resulting adversarial perturbation must deceive the model on the majority of examples:
$$
f_{\theta}(\mathcal{A}(x_i)) \neq C_i = f_{\theta}(x_i), \ \forall i = 1, \dots, N
$$

\subsection*{Multimodal attacks}
As models capable of processing multiple types of data have gained significant popularity recently, attacks targeting these models have also seen substantial growth. Since these models can handle various data modalities, the range and variety of possible attacks have increased. Moreover, these models are capable of solving multiple tasks simultaneously, unlike classical architectures, which further expands the attack surface and the potential for discovering vulnerabilities.

In vision-and-language multimodal models, there are numerous variations for modifying input data (text and images). Attacks on such models typically involve balancing the preservation of accuracy between the original and attacked data with the effectiveness and transferability of the attack. These attacks can be categorized based on the modalities they manipulate.

For altering the textual modality, there are numerous jailbreak prompts \cite{llm_att} designed to force the model to generate responses to unsafe queries. These prompts can be crafted manually \cite{like_DAN} or generated through training to produce a specific class of responses \cite{for_diff}, the latter often lacking interpretability at the language level. In many cases, a single well-chosen token is sufficient to distort the generated outputs.

Similar approaches are used for attacks on the visual modality. Since language models processing visual information are adept at recognizing content from images \cite{OCR}, in some cases, a simple text prompt overlaid on the original image is sufficient to mislead the model (\cite{from_VLM}, \cite{add_clip_loss}). For more effective results, visual tokens can be trained in the model’s latent space \cite{lat_tok} or directly within the pixel space of the original image \cite{hijacks}, \cite{VLM_UAPs}.

\section*{Method}
\subsection*{Task definition}
In a multiclass classification task, the model must learn to correctly predict classes corresponding to the given target labels. For an image \( x_0 \in \mathbb{R}^d \) and its corresponding class \( C_i \), a trained model \( f_{\theta}(x): \mathbb{R}^d \rightarrow \{1, \dots, N\} \) should predict the correct label: \( f_{\theta}(x_0) = C_i \).

\begin{figure}[h!]
    \centering
    \includegraphics[width=0.7\linewidth]{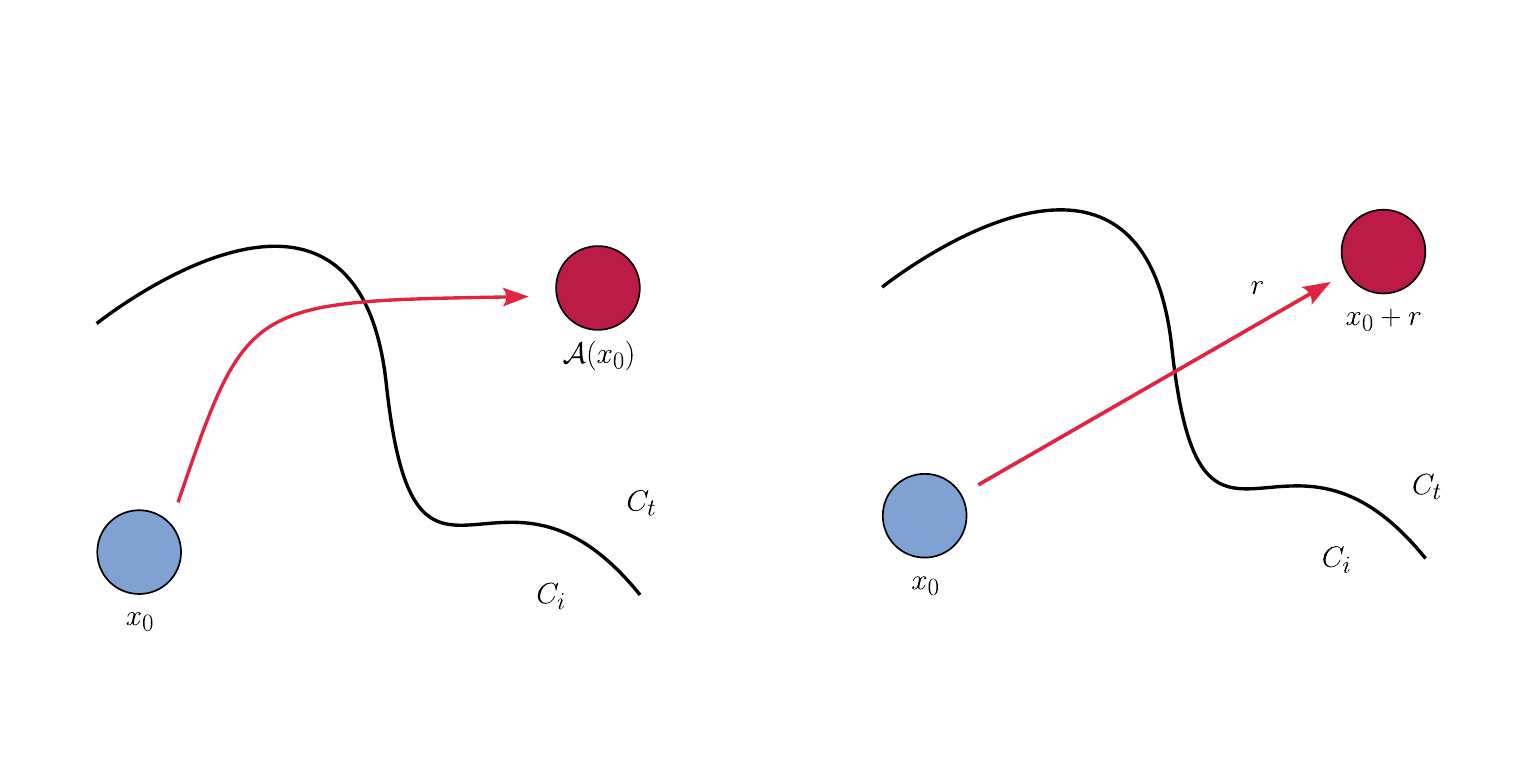}
    \caption{Model decision boundaries for clean and adversarial data in general (left) and for additive attacks (right).}
    \label{fig:attck}
\end{figure}

The attacker's objective is to choose a method of modifying the input data such that, after applying this transformation to the original images, the model predicts an incorrect or attacker-specified class (see Fig. \ref{fig:attck}). In this study, the predefined transformation \( \mathcal{A} \) involved directly altering the image by replacing part of its pixels, simulating the physical overlay of a specific sticker on the image.

\subsection*{Patch Attacks}

\begin{figure}[h!]
    \centering
    \includegraphics[width=0.5\linewidth]{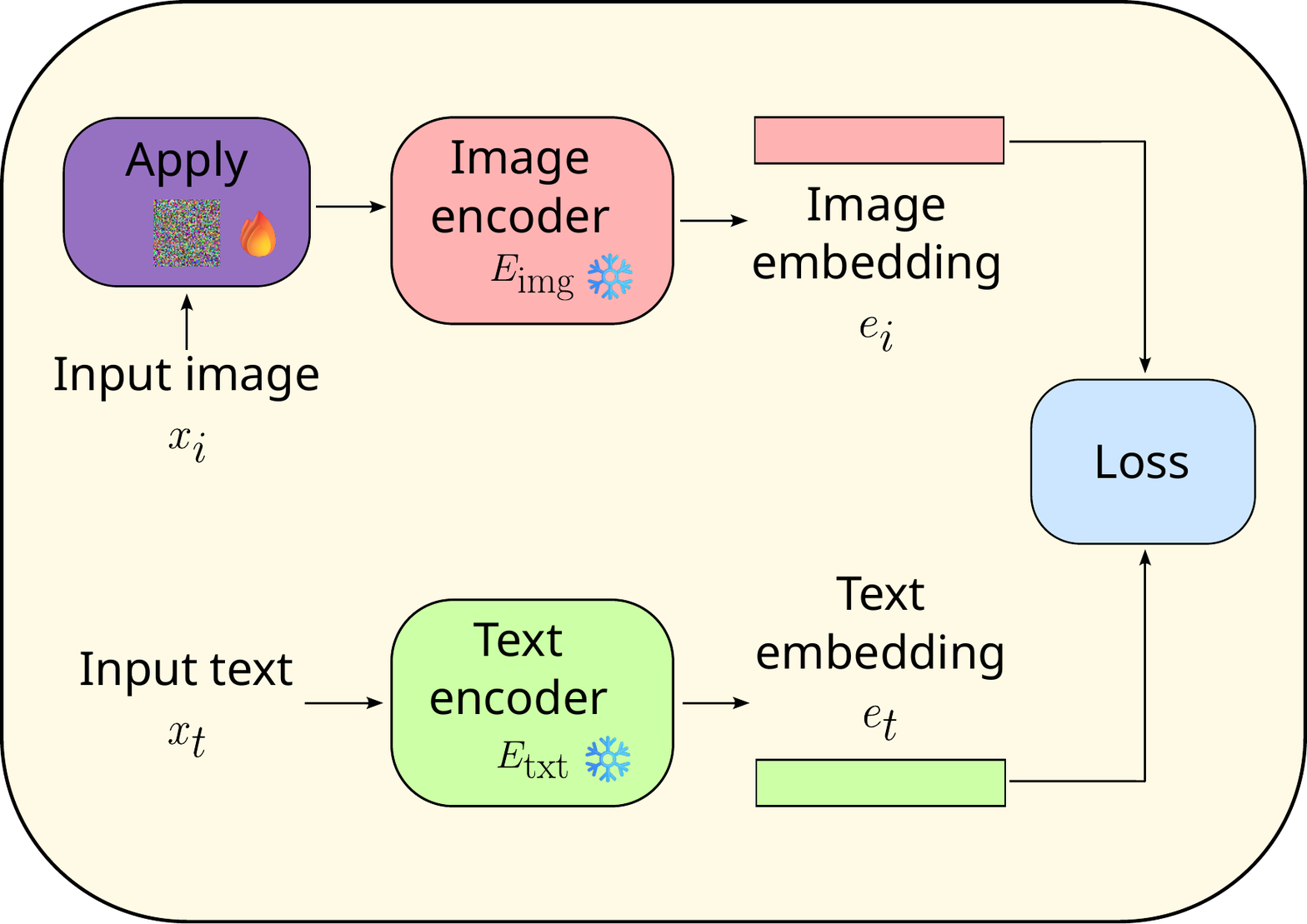}
    \caption{Pipeline for training patches to attack images in the CLIP model. A patch with trainable parameters is overlaid on the original image \( x_i \), and then embeddings \( e_i \) are calculated for the attacked images.}
    \label{fig:pipe}
\end{figure}

Let the descriptions of the images form a set \( \mathcal{T} \), and the original clean images form a set \( \mathcal{I} \). Applying a patch attack defined by \( \mathcal{A} \) to these images results in a set of adversarial images:

$$
\mathcal{I_{\text{adv}}} = \{x: x = f_{\text{prep}}(\mathcal{A}(x_0, p)) \ | \ x_0 \in \mathcal{I}\},
$$

$$
\mathcal{A}(x, p) = x \oplus p
$$

where \( f_{\text{prep}} \) is the image preprocessing function, depending on the model type \( f_{\theta} \), and \( p \) is the patch, with \( x \oplus p \) representing the operation of replacing the pixels of the original image with the patch values. In standard representation, \( f_{\text{prep}} \) includes scaling, central alignment, and normalization.

Next, embeddings for each image-text pair are computed using encoder models \( E_{\text{img}}, E_{\text{txt}} \), respectively.

$$
\mathbf{e_i} = E_{\text{img}}(I), \ I \in \mathcal{I_{\text{adv}}}
$$

$$
\mathbf{e_t} = E_{\text{txt}}(T), \ T \in \mathcal{T}
$$

For the obtained vector representations, the cross-entropy loss function is calculated:

$$
\mathcal{L_{\text{patch}}} = -\underset{(I_i, T_i) \sim \mathcal{D_{\text{train}}}}{\mathbb{E}} \left[\log \frac{e^{t \mathbf{x}_i \cdot \mathbf{y}_i}}{\sum_{j=1}^{N} e^{t \mathbf{x}_i \cdot \mathbf{y}_j}}\right]
$$

$$
\mathbf{x}_i=\frac{E_{\text{img}}\left(I_i\right)}{\left\|E_{\text{img}}\left(I_i\right)\right\|_2}, \
\mathbf{y}_i=\frac{E_{\text{txt}}\left(T_i\right)}{\left\|E_{\text{txt}}\left(T_i\right)\right\|_2}
$$

Based on the loss function, the optimization problem for the patch parameters is solved using gradient-based methods:

$$
p_{*} = \underset{p}{\text{argmin}} \ \mathcal{L_{\text{patch}}}
$$

\section*{Experiments}

\subsection*{Experimental Setup}
\textbf{Dataset:} The patches were trained on the Caltech-256 dataset \cite{Caltech}. The total number of images was about 10,000, 30\% of which were selected for validation. Additionally, two distinct collections of short videos were used for the video adaptation of the models, each containing 200 videos with a maximum duration of one minute per video. For each video, 10 representative key frames were extracted to facilitate model evaluation, focusing on capturing the essential visual content efficiently.

\noindent\textbf{Models:} For the visual encoder in CLIP, models from the OpenCLIP library \cite{open_clip} were used. There were several types of ViT (Vision Transformer), differing in size and training method: ViT-B-16, ViT-L-14, ViT-L-14-336, and ViT-B-16-SigLIP. The results of the trained patches were also tested on the multimodal model LLaVA-v1.5 \cite{llava}, which has 7 billion parameters.

\noindent\textbf{Implementation Details:}
All patches were applied to the original image before preprocessing. Each patch was trained for 5 epochs on the training data and then tested on the validation dataset. The Adam optimizer was used for the sticker parameters, with default hyperparameters and a learning rate of 0.1. To ensure more accurate results, the trained patches were applied to the same image five times during metric calculation, and the results were averaged for each image.

\noindent\textbf{Metrics:}
The main quantitative metric for attack success and transferability was the Attack Success Rate (ASR), defined as follows:

$$
\text{ASR} = \frac{N_{\text{success}}}{N_{\text{total}}} \cdot 100\% 
$$

$$
N_{\text{success}} := \{(x, C_i) \in \mathcal{D}_{\text{test}} \ | \ f_{\theta}(\mathcal{A}(x, C_i, f_{\theta}) = C_t \cap C_t \neq C_i \}
$$

$$
N_{\text{total}} := \{(x, C_i) \in \mathcal{D}_{\text{test}} \ | \ C_i \neq C_t \}
$$

where \( f_{\theta} \) is the attacked model, \( \mathcal{A} \) is the transformation, and \( \mathcal{D}_{\text{test}} \) is the validation dataset. \( C_t \) is the target label set by the attacker.

Additionally, a variation called ASR@5 was used, where instead of using the top-1 label for \( N_{\text{success}} \), the top-5 predicted labels for each attacked image were checked. ASR is a key indicator that measures the proportion of successful adversarial attacks out of the total number of attacks. A higher ASR implies good transferability of the adversarial examples, indicating the attack's effectiveness in compromising the model under various conditions.

For multimodal models like LLaVA, perplexity was also calculated, which is defined as the exponent of the mean negative log-likelihood of a sequence. Given a sequence of generated tokens \( X=\left(x_0, x_1, \ldots, x_t\right) \), perplexity \( \text{PPL}(X) \) is computed as follows:

$$
\operatorname{PPL}(X)=\exp \left(-\frac{1}{t} \sum_{i=1}^t \log p_\theta\left(x_i \mid x_{<i}\right)\right)
$$

where \( \log p_\theta\left(x_i \mid x_{<i}\right) \) is the log probability of the i-th token given all previous tokens \( x_{<i} \). This is also equivalent to the exponent of the cross-entropy between the data and the model's predictions.

\subsection*{The results of applying patches to images}
First, patches of different sizes were trained for the ViT-B-16 model. Their generalization ability was also tested on other variations of the image encoder. To measure the effect of size, square-shaped patches with widths and lengths of 64 pixels, as well as patches sized 32 pixels, were used as the primary sizes. For trial classes, the labels "cake" and "homer-simpson" from the Caltech-256 dataset were chosen.

The results are presented in Table \ref{vit-b-res}. The trained patches themselves can be viewed in the supplementary materials (see Figures \ref{patchas_ViT_B_n_forms}, \ref{patches_ViT_Ls}). Two metrics, ASR@1 and ASR@5, were measured to assess the success of the trained patches. For simple patches (column "w/o text"), the results show that for the first metric, patches of size 32 pixels yield a success rate of around 40\%, while patches twice as large achieve results around 70-90\% when looking at the models for which these patches were trained (bolded results). In almost all cases, the top-5 labels for the attacked images reach an approximately 90\% success rate. Adding text under the patch during training and validation (column "with text") does not significantly affect the final metrics.

It is also worth noting that there is a non-zero transferability of this type of attack to models of similar or even larger sizes, specifically from the ViT-B-16 model (for which the sticker was trained) to the ViT-L-14 and SigLIP-B-16 models.

Additionally, for comparison, metrics were measured on two additional types of images: clean images (row "clean images") and images with an additional cake sticker (row "control cake"). Comparing these with the previous results, it's clear that patch attacks show significantly better results, especially for the top-1 labels, while on clean images the model makes almost no mistakes.

\begin{table}
    \centering
    \begin{tabular}{c c c c c c c |c}
        \hline
        \multirow{3}{*}{class} & \multirow{3}{*}{Metrics} & \multicolumn{5}{c|}{w/o text} & \multicolumn{1}{c}{w/ text} \\ \cline{3-8}
        & & \multicolumn{1}{c|}{32x32} & \multicolumn{5}{|c}{64x64} \\ \cline{3-8}
        & & \textbf{B-16} & \textbf{B-16} & L-14 & \textbf{L-14-336} & SigLIP-B-16 & \textbf{B-16} \\ \hline
        
        \multirow{2}{*}{cake} & ASR@1 & 0.4851 & 0.8112 & 0.3936 & 0.6915 & 0.2073 & 0.8062 \\ 
        & ASR@5 & 0.7295 & 0.9078 & 0.8844 & 0.9316 & 0.8865 & 0.9081 \\ \hline
        
        homer- & ASR@1 & 0.4131 & 0.8466 & 0.6529 & 0.8943 & 0.3503 & 0.7689 \\ 
        simpson & ASR@5 & 0.6468 & 0.9210 & 0.9206 & 0.9469 & 0.8219 & 0.8822 \\ \hline\hline
        
        Control & ASR@1 & 0.0565 & 0.1752 & 0.1707 & 0.1626 & 0.1511 & 0.4932 \\
        cake & ASR@5 & 0.3450 & 0.7479 & 0.8544 & 0.8687 & 0.8816 & 0.8547 \\ \hline
        
        Clean & ASR@1 & 0 & 0 & 0 & 0 & 0 & 0 \\
        images & ASR@5 & 0.0004 & 0.0004 & 0.0003 & 0.0002 & 0.0008 & 0.0004 \\ \hline
    
    \end{tabular}
    \caption{Performance for various configurations of the ViT model. The models on which the stickers were trained are highlighted in bold.}
    \label{vit-b-res}
\end{table}

The generated patches often display some meaningful parts of the target labels (see Section A of the supplementary materials). Therefore, Table \ref{another_type_att} presents results for non-trainable patches with text (columns “text patch”), where the text size was adjusted for each image so that the final rectangular patch covered \(a\) percent of the original clean image before preprocessing. Results were also obtained for patches in the form of a frame, with the width \(W\) manually chosen.

These results show, firstly, that the model pays significant attention to text. As a result, during patch generation, meaningful phrases appear that resemble the target label, helping the model. Secondly, even a frame patch with a width of just 4 pixels yields results on par with or even better than patches sized 64×64 pixels.

\begin{table}
    \centering
    \begin{tabular}{c c c c c c | c c c c}
        \hline
        \multirow{2}{*}{class} & \multirow{2}{*}{Metrics} & \multicolumn{4}{c|}{text patch} & \multicolumn{4}{c}{frame patch} \\ \cline{3-10}
        & & a = 0.1 & a = 0.2 & a = 0.4 & a = 0.6 & W = 1 & W = 2 & W = 3 & W = 4 \\ \hline
        \multirow{2}{*}{cake} & ASR@1 & 0.1864 & 0.3798 & 0.6106 & 0.8340 & 0.0030 & 0.2476 & 0.7294 & 0.9226 \\
        & ASR@5 & 0.6832 & 0.8432 & 0.9421 & 0.9818 & 0.0708 & 0.5212 & 0.9164 & 0.9636 \\ \hline
        homer- & ASR@1 & 0.0003 & 0.1701 & 0.6394 & 0.8756 & 0.0030 & 0.3919 & 0.7114 & 0.8701 \\
        simpson & ASR@5 & 0.0054 & 0.4687 & 0.9479 & 0.9934 & 0.0764 & 0.6674 & 0.8537 & 0.9400 \\ \hline
    \end{tabular}
    \caption{Performance for various patch values in the form of a text of size $a$ and a patch in the form of a frame of width $W$. All results were calculated for the ViT-B-16 model.}
    \label{another_type_att}
\end{table}


Similarly, results were obtained with the classic training approach for a larger number of classes (see Section A of the supplementary materials), trained for the ViT-L-14 model and tested on the smaller ViT-B-16 model. The final results are shown in Figure \ref{L-14}. It can be seen that for the L-type model, ASR@1 and ASR@5 average around 70\% and 85\%, respectively. For the B-type model, the results are around 65\% and 30\%, respectively. However, in both cases, there are dips in performance, which correspond to so-called “hard classes”. These are classes where the target label represents a generalization of a certain object, concept, or phenomenon. This could be due to the fact that during the initial training of the attacked model, there were too many variations of images for these hard classes, preventing the model from encoding all the information into much smaller patch sizes.



\subsubsection*{Noise augmentations}
In Figure \ref{L-14} of supplementary materials, section A.1., as previously noted, hard classes are also shown, where the results were very low, especially when transferred to another model. To improve the success rate of attacks without access to the target model, patches can be trained with augmentations. This approach may affect the final result for the model the patch was trained on but will improve transferability.

However, not every type of augmentation can improve the metric. To select a set of such applicable changes, the augmentation function should reduce interactions between different regions of the patch (\cite{aug_UAP}). A suitable function for this is the classic RandomCrop method, which cuts a random rectangular section from the original patch. Therefore, the following method was used to update the patch:

$$
p \leftarrow p + \beta \cdot (\text{R} \circ \text{RC})(p)
$$
$$
x_{\text{adv}} \leftarrow f_{\text{prep}}(\mathcal{A}(x_0, p)), \ x_0 \in \mathcal{I}
$$

where \(\text{RC}\) represents the RandomCrop operation, and \(\text{R}\) is the Resize operation to resize the patch back to its original size. The default value for \(\beta\) was set to 1. This separate enhancement of the usefulness of local regions of the patch contributes to the overall effectiveness of the Universal Adversarial Perturbation (UAP).

\begin{table}
    \centering
    \begin{tabular}{c c c c c c c c c c}
        \hline
        \multirow{2}{*}{Model} & \multirow{2}{*}{C\&R} & \multirow{2}{*}{Metric} & \multicolumn{7}{c}{class} \\ \cline{4-9}
        & & & rainbow & saturn & people & galaxy & clutter & fireworks  \\ \hline
        
        \multirow{4}{*}{L-14} & \textcolor{green}{\ding{51}} & \multirow{2}{*}{ASR@1} & 0.3095 & 0.7643 & 0.1366 & 0.8031 & 0.3755 & 0.5151 \\
        
        & \textcolor{red}{\ding{55}} &  &
        \colorbox{lightgray}{0.5050} &
        \colorbox{lightgray}{0.8386} & \colorbox{lightgray}{0.4795} & \colorbox{lightgray}{0.8260} & \colorbox{lightgray}{0.7419} & \colorbox{lightgray}{0.7194} & \\ \cline{2-9}
        
        & \textcolor{green}{\ding{51}} & \multirow{2}{*}{ASR@5} & 0.8804 & 0.8967 & \colorbox{lightgray}{0.7852} & 0.9171 & 0.7946 & 0.8855 \\
        & \textcolor{red}{\ding{55}} &  & \colorbox{lightgray}{0.8839} & \colorbox{lightgray}{0.9120} & 0.7828 & \colorbox{lightgray}{0.9251} & \colorbox{lightgray}{0.8764} & \colorbox{lightgray}{0.9009} \\ \hline\hline
        
        \multirow{4}{*}{B-16} & \textcolor{green}{\ding{51}} & \multirow{2}{*}{ASR@1} & \textbf{0.1607} & \textbf{0.3535} & \textbf{0.2331} & \textbf{0.2194} & 0.0852 & \textbf{0.3391} \\
        
        & \textcolor{red}{\ding{55}} & & 0.0401 & 0.1766 & 0.0096 & 0.2003 & \textbf{0.1911} & 0.1514 \\ \cline{2-9}
        
        & \textcolor{green}{\ding{51}} & \multirow{2}{*}{ASR@5} & \textbf{0.7322} & \textbf{0.7966} & \textbf{0.7337} & 0.7705 & 0.5122 & \textbf{0.8161} \\
        
        & \textcolor{red}{\ding{55}} &  & 0.3979 & 0.6864 & 0.1850 & \textbf{0.8064} & \textbf{0.5503} & 0.5945 \\ \hline
        
    \end{tabular}
    \caption{Comparison of attack success metrics for two models of different sizes (ViT-B-16, ViT-L-14) in the presence and absence of noise augmentation during training. Patches were trained for the ViT-L-14 model. Gray highlights the maxima among the metrics ASR@1, ASR@5 for the ViT-L-14 model, and bold — for the ViT-B-16 model, respectively}
    \label{cr_mode}
\end{table}

Table \ref{cr_mode} shows the results of training patches with the updated method, adding the RandomCrop and Resize augmentations for some target labels from the set of hard classes (which previously had the lowest results). The obtained data indicate that, on average, the transfer results with augmentations improved several times over. For some of these classes, the result increased from almost zero to 20-50\% for the ASR@1 metric. Similarly, the ASR@5 metric improved for nearly all the selected classes. However, due to this high performance, the results for the model on which the patches were trained—in this case, ViT-L-14—became worse than they would have been without augmentations.

\begin{figure}[h!]
    \centering
    \includegraphics[width=0.73\linewidth]{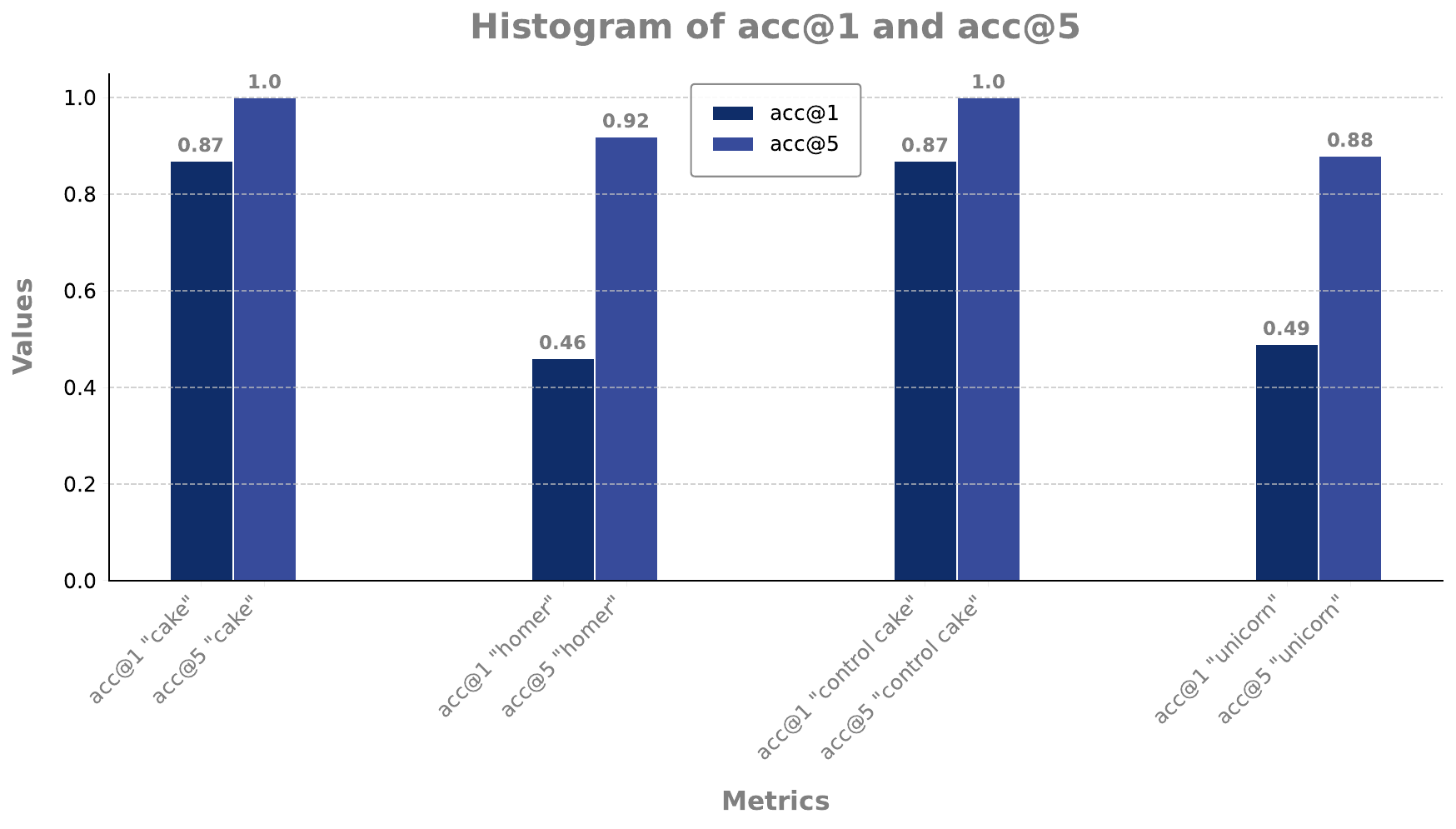}
    \caption{A diagram of the occurrences of the target class in the model response. For patches with the image of a real image of a pie, both metrics turned out to be equal to 0.}
    \label{llava_hist}
\end{figure}

\subsection*{The result for LLaVA-1.5 model}
In multimodal models, it is common to use pre-trained components, so one can attempt to attack only a part of a complex model and assess the success of such an attack. This section explores the result of an attack on the LLAVA-1.5-7b model with a ViT-L-14-336 image encoder. For this model, patches of various classes were trained: 'cake', 'homer-simpson', 'unicorn', and the result was measured on a set of 25 images. Each patch was randomly applied five times to an image, and for these subsets, the clean image and its five attacked variations were fed into the model to generate a response for each.

\begin{figure}[h!]
    \centering
    \includegraphics[width=0.65\linewidth]{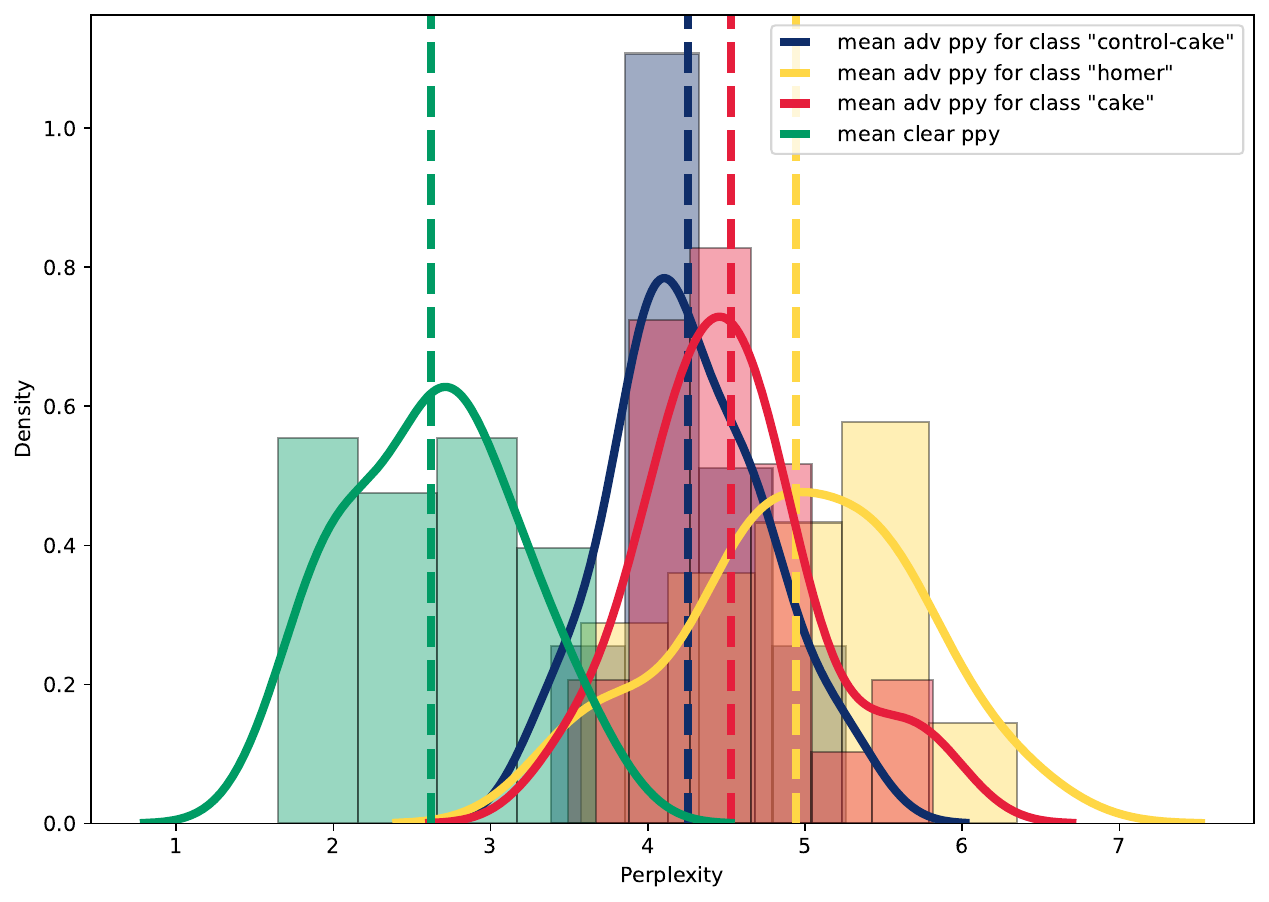}
    \caption{Perplexy for selected attacked images.}
    \label{PPY_llava_1.5}
\end{figure}

From the graph in Figure \ref{PPY_llava_1.5}, it can be seen that the model hesitates significantly in its response for attacked images compared to the original ones. It is also evident that for an attack using a simple image of a cake, the model's perplexity is lower. For specific model responses with particular images, refer to section B.2. of the supplementary materials.

Also, the presence of the target class was checked in the model's responses. The results are shown in the bar chart (see Figure \ref{llava_hist}). It displays results for several classes, as well as for a patch with random initialization values. For all results, the top-1 accuracy remains around 90\%. It is worth noting that for the 'cake' class, the results were the same as for the control variation of the patch, although the model's responses were more consistent: they did not include a simple enumeration of objects (see section G of the supplementary materials).

\subsection*{The results of applying patches to videos}
In the simplest case, a video can be represented as a set of frames. Therefore, for most tasks, video processing can be approached as simultaneous processing of multiple frames. In retrieval tasks for video content, a similar approach can be applied, meaning that image attacks can be transferred to video as well. This section examines patch attacks for video frame sequences based on previously trained patches.

First, let's formally define a patch attack for video. Let us have a dataset \(\mathcal{V}\) consisting of various sets of videos. Then, for each video \(v \in \mathbb{R}^{T\times C \times W \times H}\), we first select the so-called key frames \(v_{\text{key}} \in \mathbb{R}^{T^{\prime}\times C \times W \times H}\), based on some heuristics or trained algorithms, ensuring that \(T^{\prime} \ll T\). The dataset of such key frames for each video is denoted as \(\mathcal{I_{\text{key}}}\). An algorithm, the pseudocode for which is shown in supplementary section B.1., was applied to these frames.







\begin{figure}[h!]
    \centering
    \includegraphics[width=0.7\linewidth]{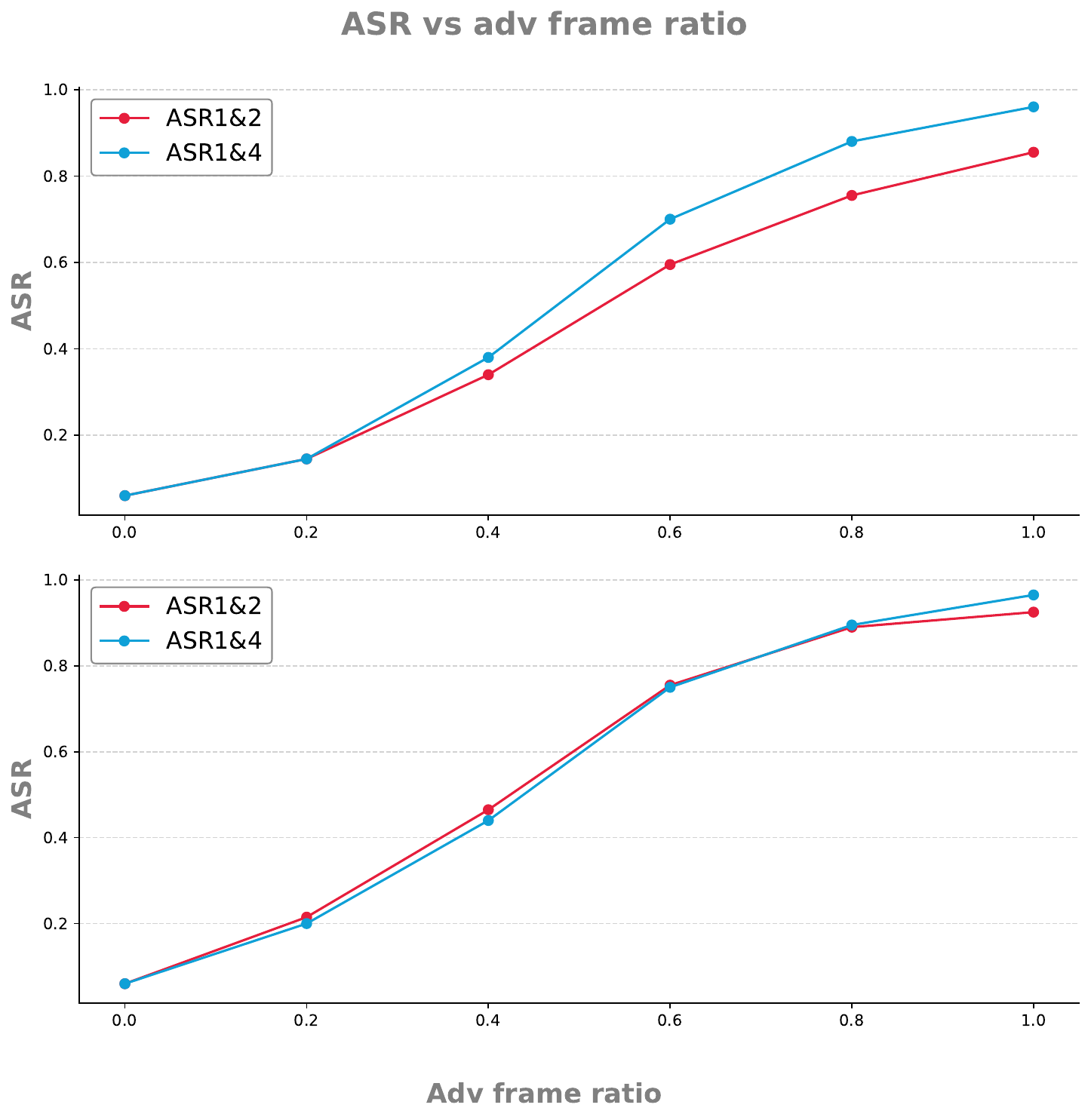}
    \caption{The metric of attack success depends on the proportion of attacked video frames. For the Dzen dataset (top) and for the TikTok dataset (bottom). ASR1\&2 and ASR1\&4 denote the ratio of CLIP Score 1 exceeds CLIP Score 2 or CLIP Score 4 over given video datasets, respectively.}
    \label{adv_vid_cake}
\end{figure}

\begin{figure}[h!]
    \centering
    \includegraphics[width=0.9\linewidth]{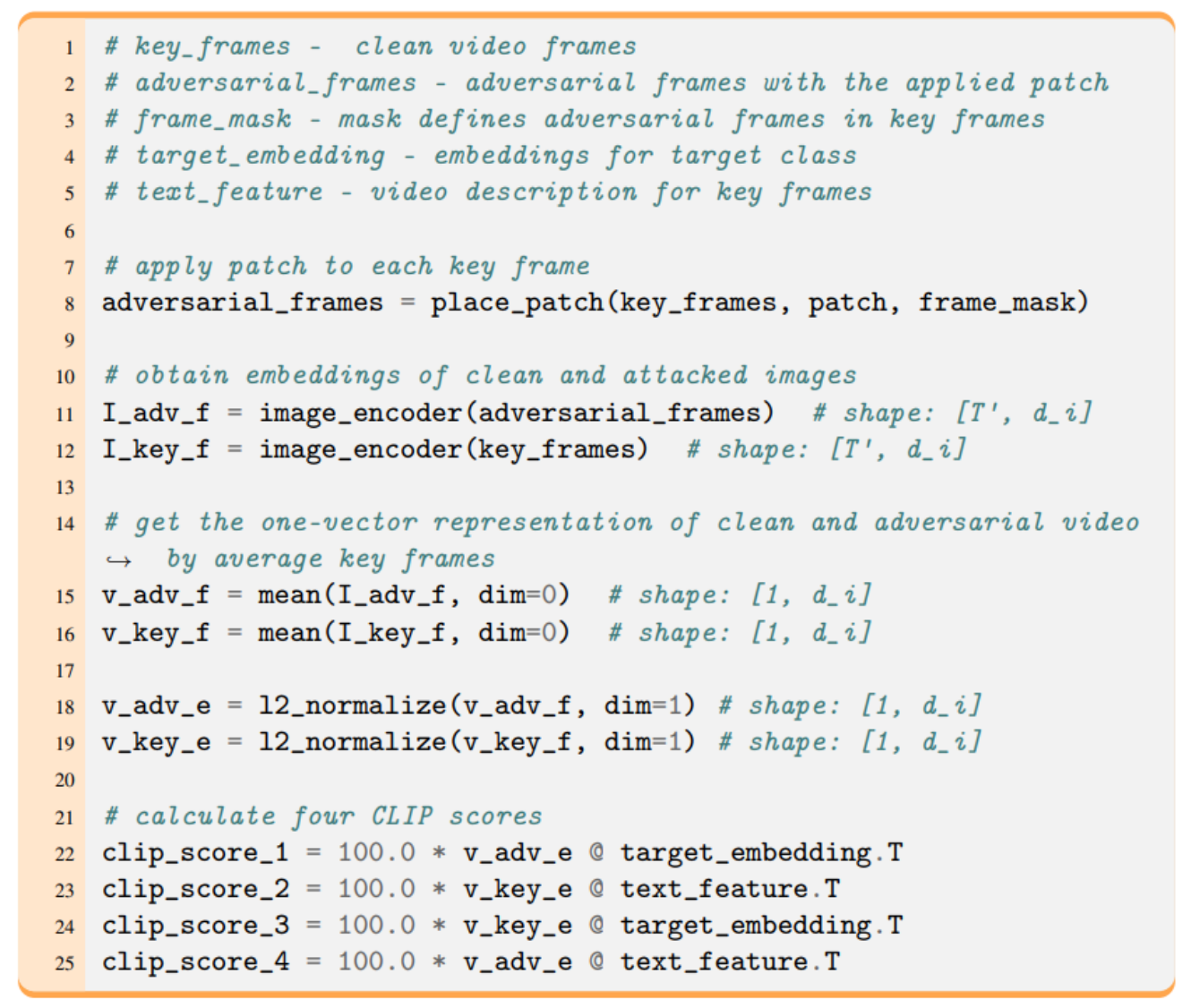}
    
    \caption{Algorithm of calculating CLIP scores for video frames}
    \label{vid_algo}
\end{figure}

Applying such algorithm to the \(\mathcal{I_{\text{key}}}\) dataset, four CLIP Score values were calculated. One for clean images and their original captions (CLIP Score 2), another for attacked images and captions in the form of the target label (CLIP Score 1), and two others where only one of the modalities was changed (CLIP Score 3, CLIP Score 4), which shows the model's confidence in such cases (see Figure \ref{vid_algo}).

An attack on a video search engine, represented by the Multimodal-CLIP-ViT-L-14 model, was considered successful if CLIP Score 1 exceeded CLIP Score 4 for a specific video (variant 1). In this case, the focus is on how much the attacked video is more relevant to the target class than to the original caption. Additionally, another similar result was considered, but for CLIP Score 1 and CLIP Score 2, respectively (variant 2). This showed how relevant the attacked images were to the target label compared to the clean frames and their descriptions.

The results were measured on video datasets from Dzen and TikTok, each containing 10 frames. Some frames were attacked, while the others were left untouched. The success rate of the attack was measured depending on the proportion of attacked frames. The final results for the patch with the specified class 'cake' are presented in Figure \ref{adv_vid_cake}. The resulting ASR dependency almost linearly matches the proportion of attacked frames. Therefore, the more the model needs to be confused, the more attacked frames should be included in the original video.

\begin{figure}[h]
    \centering
    \includegraphics[width=0.9\linewidth]{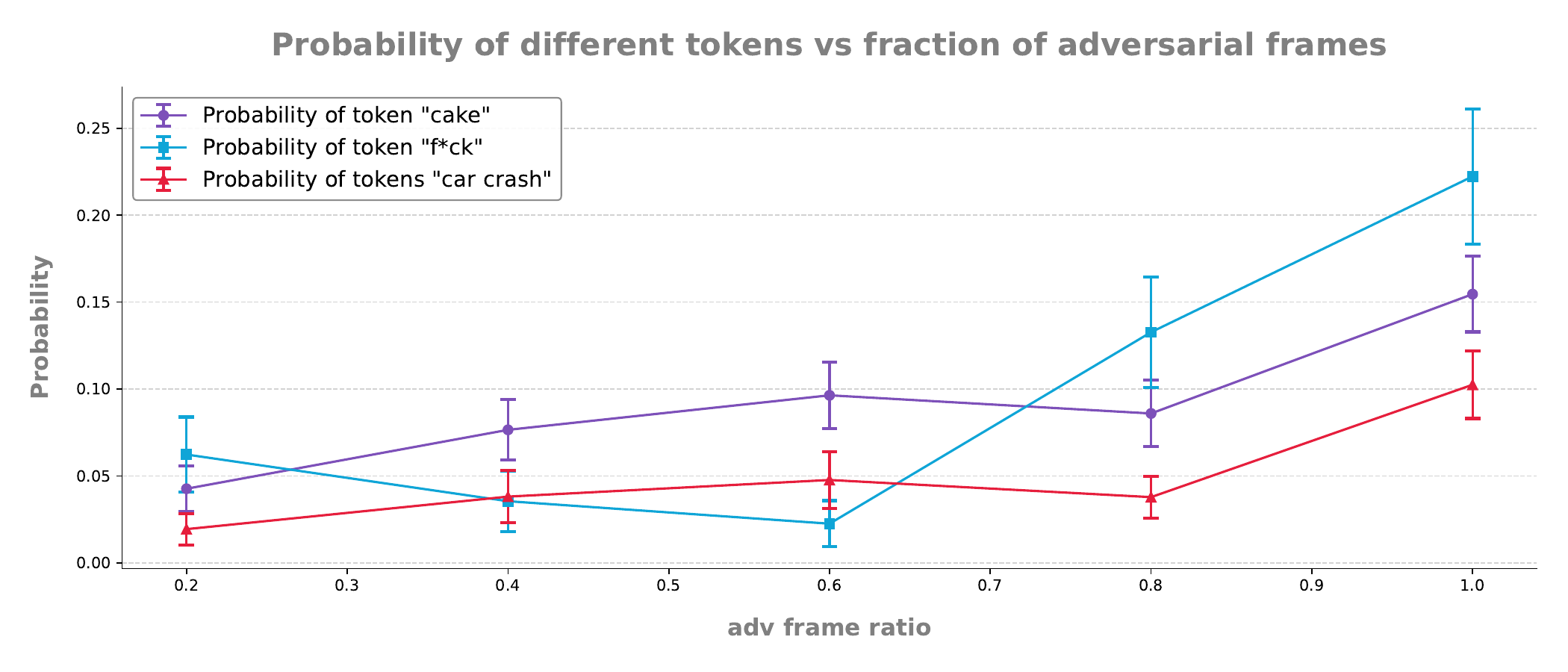}
    \caption{Probability for adv-tokens for LLaVA-OneVision 0.5b model}
    \label{llava-vids_1}
\end{figure}

\subsection*{The results for LLaVA-OneVision model}

\begin{figure}
    \centering
    \includegraphics[width=0.9\linewidth]{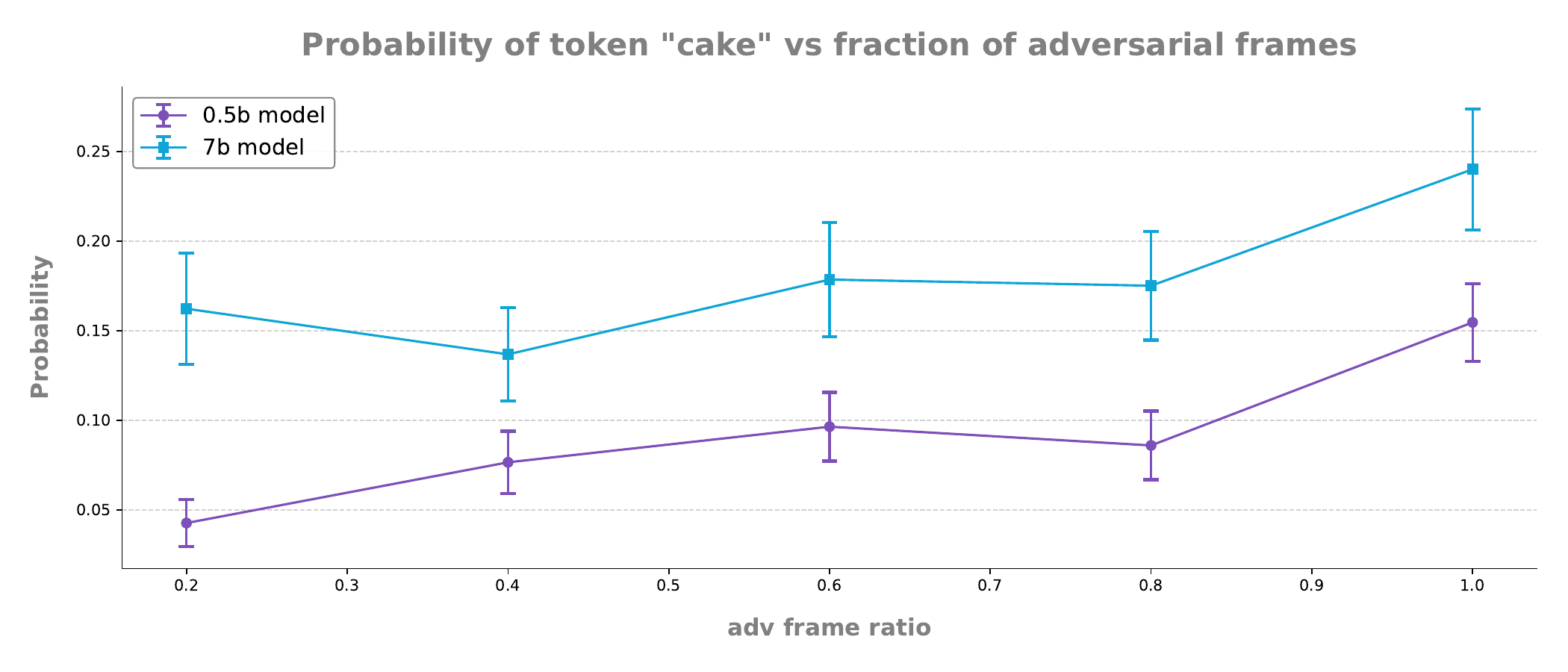}
    \caption{Comparison with 0.5b and 7b models}
    \label{llava-vids_2}
\end{figure}

LLaVa-OneVision (\cite{llava-vid}) is a more modern model, a version of LLaVA but for video, capable of processing and answering questions related to them, was also tested. Similarly, it takes key frames as input and operates on them along with a user-provided text prompt. In all experiments, it was a question: "What is shown in this video?". The model's response included text as well as probabilities for each token in the final answer. This procedure was carried out on the Dzen dataset with varying degrees of attacked pictures in the key frames. The resulting dynamics depending on the number of attacked frames are shown in Figure \ref{llava-vids_1}, \ref{llava-vids_2}. The same results but for different model size numbers are shown on the second part of this image. Interestingly, that for bigger model, its response with the adversarial token is more probably than for smaller one. It can be observed that there is a positive correlation with the number of attacked frames in the original video. Although the final result is not as significant as for image-based models, it is still non-zero.

\section*{Conclusion} In this work, we conducted a study of patch attack variations on CLIP models of different sizes, as well as on the LLaVA and LLaVA-OneVision models, to assess the vulnerabilities of multimodal language models. Even with a small patches of size 64x64 the adversary can achieve up to 90\% success in average for classification models, like CLIP. Furthermore, for LLaVA family of models, particularly trained patches for ViT backbone can discourage the whole model and poison the final responses, which include in general the target label in 60\% of cases. Additionally, there are interesting results for video models: for simple CLIP embeddings, the data shows that more adversarial frames can lead to better results in terms of success metric; for OneVision version of video model simple patches also can alter the model's answer in the best cases showing 25\% of success.

Thus, these models do indeed inherit vulnerabilities from their component parts, especially those that are openly accessible. The analysis showed that model size plays a significant role in attack resilience, confirming the need for careful selection and testing of model components before they are used in real-world applications and systems that may interact with external users.

The results of the study highlight the importance of continued work in the area of multimodal language model security, given their growing popularity and integration across various industries. It is important to note that using open-source models in their original form in production environments carries significant risks. To ensure the reliability and security of such systems, they must be further adapted and fine-tuned. This process not only improves model quality but also enhances their resilience to attacks. In the future, researchers and developers should focus more on identifying and eliminating potential vulnerabilities to ensure the safe operation of such systems.

\section*{Acknowledgements} The research was carried out using the supercomputer complex of the National Research University Higher School of Economics \cite{cHARISMa}, for which the authors express their gratitude.

\newpage

\newpage
\section*{Supplementary materials.}
\subsection*{A. Images of trained patches}
\label{patch_examples}

\begin{figure}[h!]
    \centering
    \includegraphics[width=0.95\linewidth]{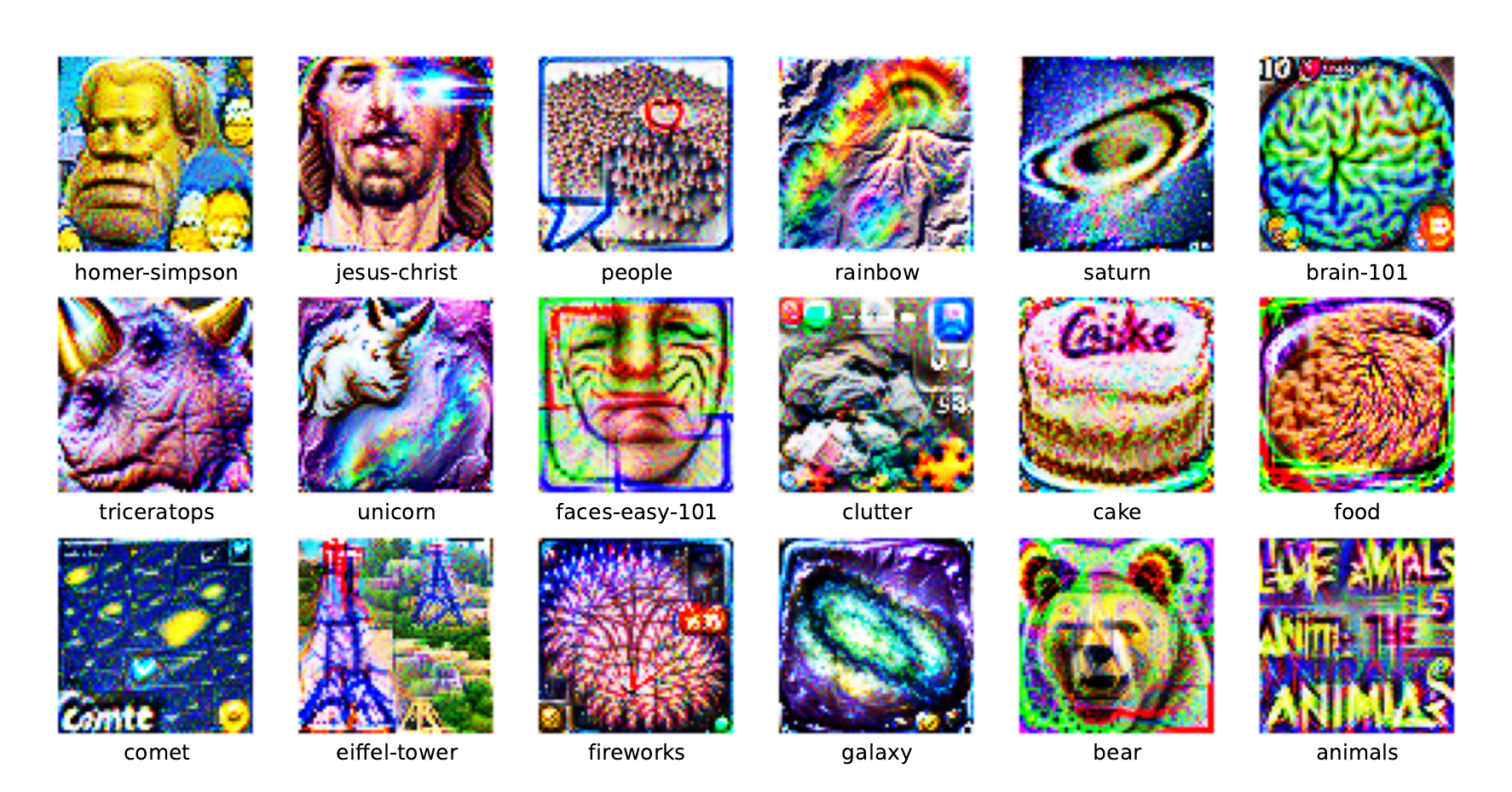}
    \caption{$64\times 64$ trained patches for the ViT-L-14 model.}
    \label{patches_ViT_Ls}
\end{figure}

\begin{figure}[h!]
    \centering
    \includegraphics[width=0.6\linewidth]{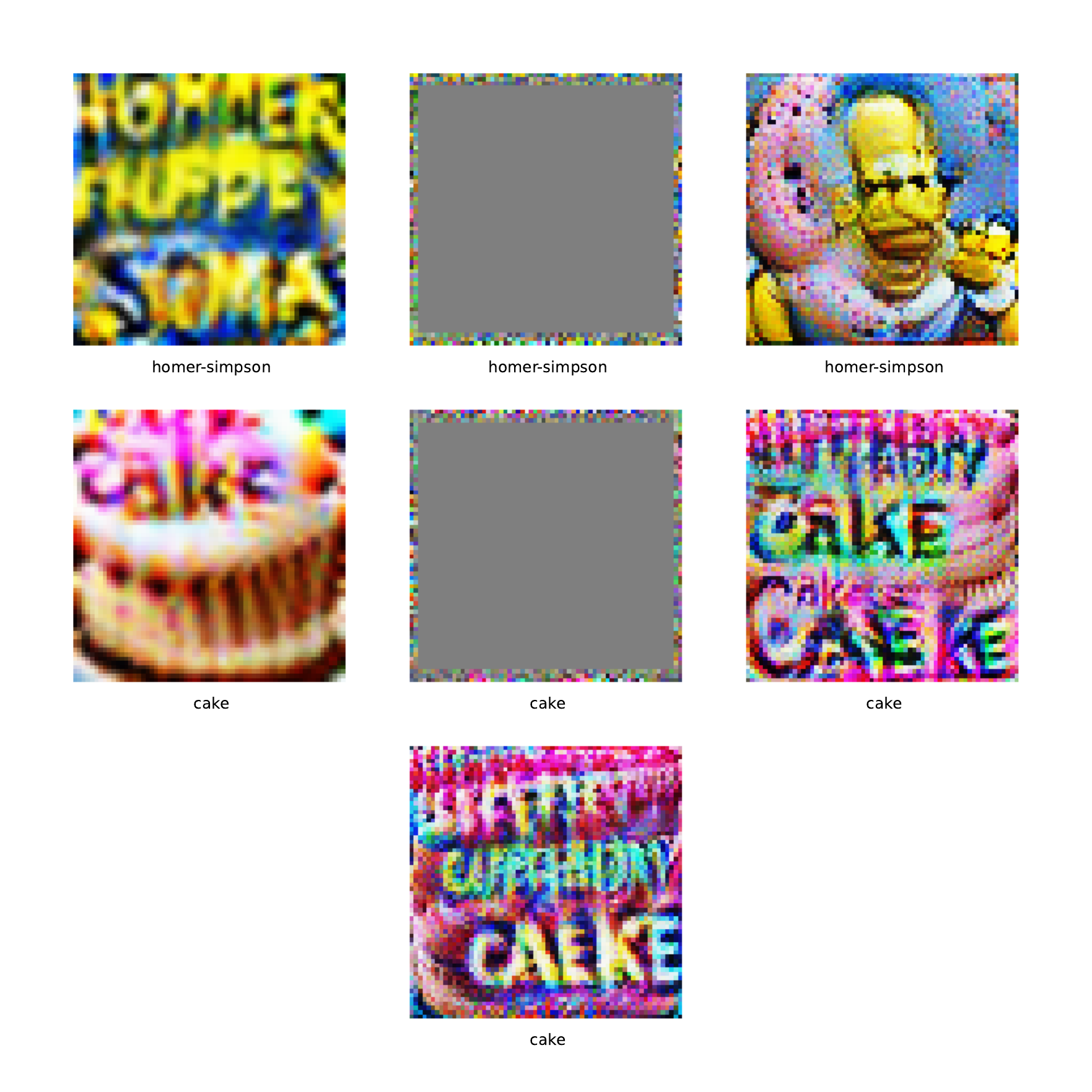}
    \caption{Trained patches for the ViT-B-16 model in different shapes and sizes. For ease of visualization, all stickers are shown in a size of 64 by 64 pixels.}
    \label{patchas_ViT_B_n_forms}
\end{figure}

The image \ref{patches_ViT_Ls} shows $64\times 64$ trained patches for the ViT-L-14 model. They clearly show inscriptions that partially correspond to the specified classes on some stickers. The text on smaller stickers and models with fewer parameters is even more visible (see Fig. \ref{patchas_ViT_B_n_forms}). As you can see, for such a model it is easier to write text, and even understandable to a person, than to generate something meaningful from the point of view of visual context.

\subsubsection*{A.1. The results of applying pathces for different target labels and model sizes}
\begin{figure}[h!]
    \centering
    \includegraphics[width=\linewidth]{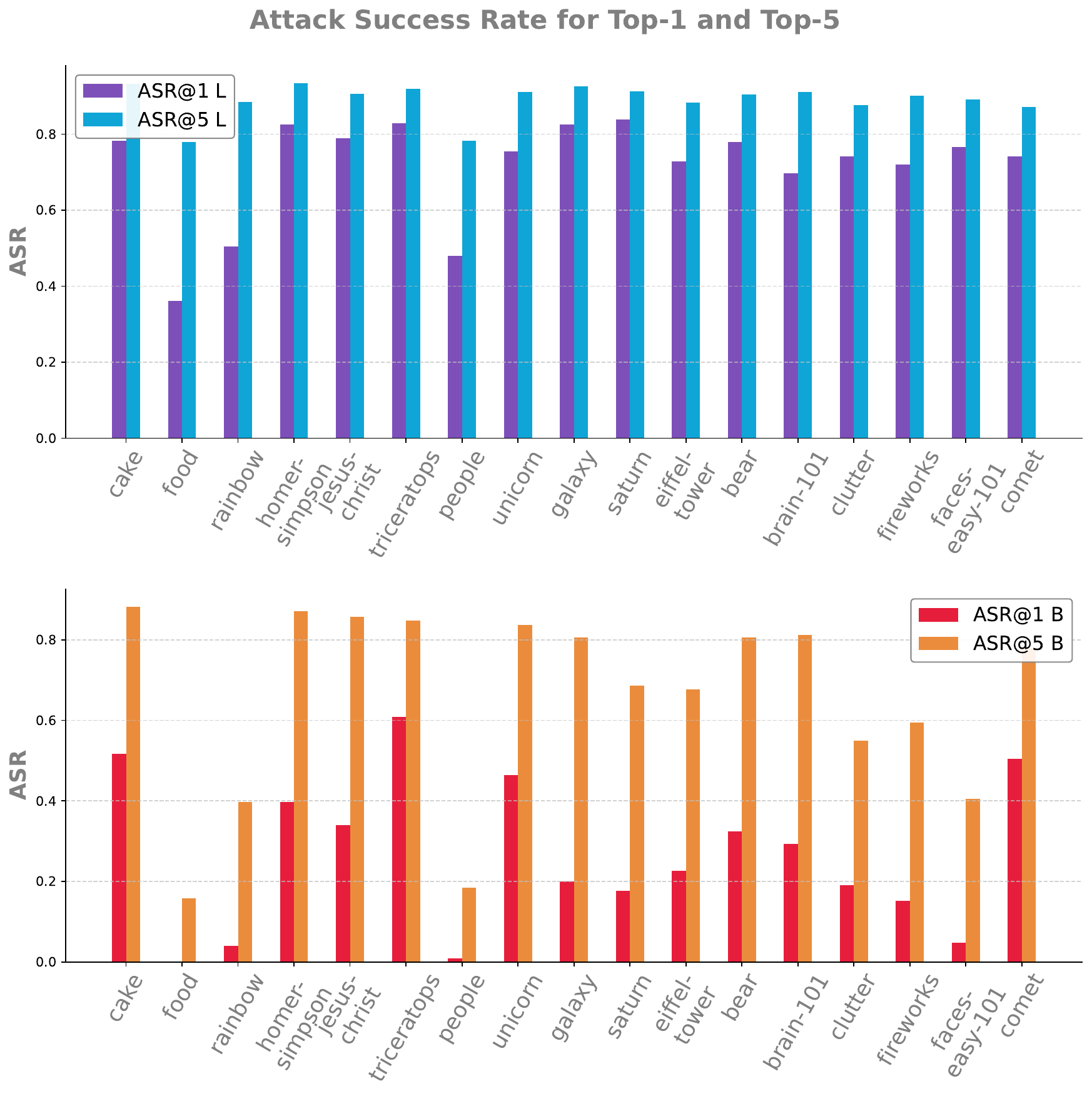}
    \caption{The attack success metric for various classes for the ViT-L-14 (top) and ViT-B-16 (bottom) models. The $64\times64$ patches themselves were trained for the ViT-L-14 model.}
    \label{L-14}
\end{figure}

The bar charts in figure \ref{L-14} show the data of success rate metric for different classes, on which pathces were trained.

\subsection*{B. CLIP Score metrics for videos.}
\label{video_metrics}
\begin{figure}[h!]
    \centering
    \includegraphics[width=0.73\textwidth]{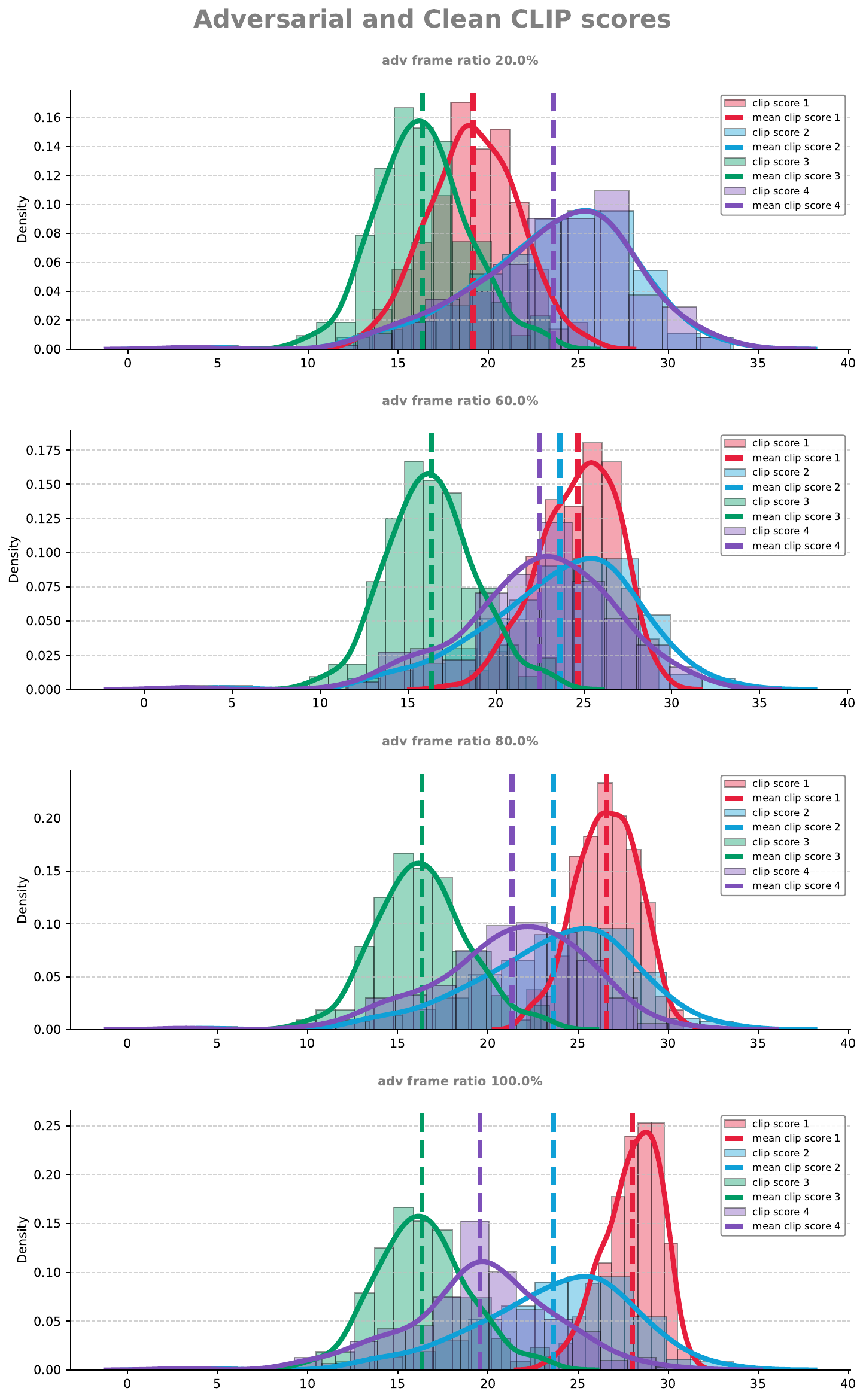}
    \caption{CLIP Score values for different degrees of attacked images in each video from the Dzen dataset.}
    \label{dzen_cake}
\end{figure}

\begin{figure}[h!]
    \centering
    \includegraphics[width=0.73\textwidth]{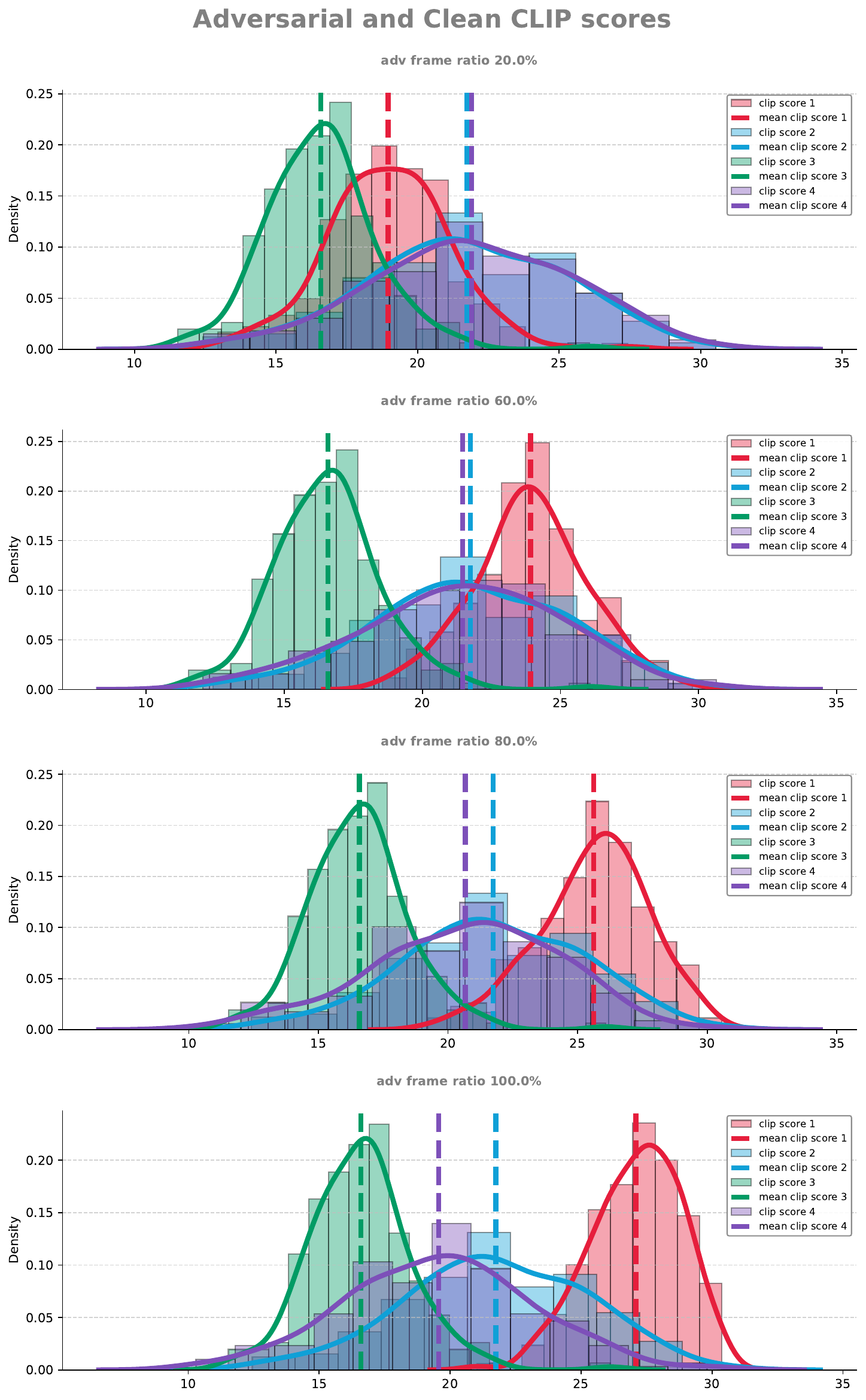}
    \caption{CLIP Score values for different degrees of attacked images in each video from the TikTok dataset.}
    \label{tiktok_cake}
\end{figure}

The graphs \ref{dzen_cake}, \ref{tiktok_cake} show the values of various types of CLIP Score metrics at different degrees of frame infection. It can be seen that for videos from the TikTok dataset, the values of the ratings are much noisier than for the Dzen dataset. And the more noisy the main frames are, the higher the CLIP Score 1 becomes compared to other estimates.

\clearpage
\newpage
\clearpage

\subsubsection*{B.1. Code for calculation of CLIP scores}\label{B.1.}
The code below shows the basic algorithm of calculation of CLIP Scores for video adversarial attack. It uses clean key frames, adversarial frames, which were obtained by applying patches to them, and descriptions of the videos with the target label of an applying patch.

\subsubsection*{B.2. Examples of answers from the LLaVA-1.5-7b and LLaVA-OneVision-7b models.}
\label{B.2.}

\begin{figure}[h!]
    \begin{tcolorbox}[colback=blue!5!white,
      colframe=blue!70!black,
      boxrule=0.5mm,
      width=\textwidth,
      arc=4mm,
      auto outer arc,
      boxsep=1mm,
      before skip=3mm,
      after skip=3mm]
        \begin{minipage}{0.24\textwidth}
            \centering
            \includegraphics[width=1.8\linewidth]{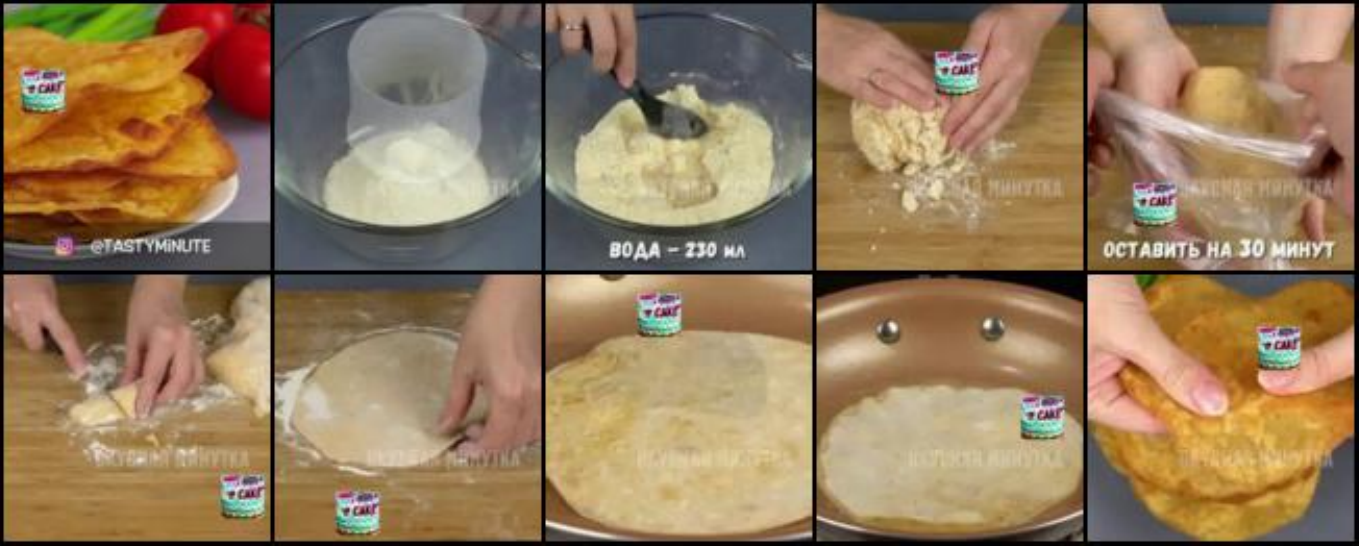}
        \end{minipage}%
        \hfill
        \begin{minipage}{0.52\textwidth}
            \textbf{User:} What is shown in this video? \\
            \\
            \textbf{Model:} The video shows \textcolor{red}{the process of making a cake}, including mixing ingredients, kneading dough, and frying the dough to create crispy, golden-brown pieces.
        \end{minipage}
    \end{tcolorbox}
\end{figure}

\newpage
\clearpage
\begin{figure}[h!]
    \begin{tcolorbox}[colback=blue!5!white,
      colframe=blue!70!black,
      boxrule=0.5mm,
      width=\textwidth,
      arc=3mm,
      auto outer arc,
      boxsep=1mm,
      before skip=5mm,
      after skip=5mm]
        \begin{minipage}{0.24\textwidth}
            \centering
            \includegraphics[width=1.8\linewidth]{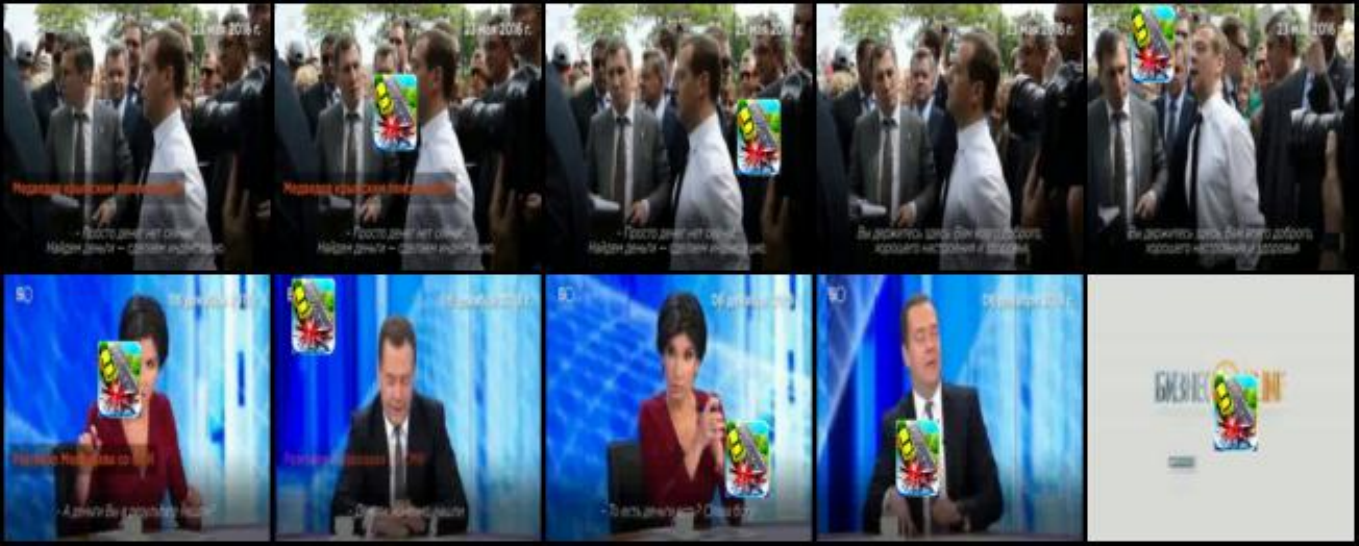}
        \end{minipage}%
        \hfill
        \begin{minipage}{0.52\textwidth}
            \textbf{User:} What is shown in this video? \\
            \\
            \textbf{Model:} The video shows a news segment discussing \textcolor{red}{a traffic accident involving a yellow car.}
        \end{minipage}
    \end{tcolorbox}
\end{figure}

\begin{figure}[h!]
    \begin{tcolorbox}[colback=blue!5!white,
      colframe=blue!75!black,
      boxrule=0.5mm,
      width=\textwidth,
      arc=3mm,
      auto outer arc,
      boxsep=1mm,
      before skip=5mm,
      after skip=5mm]
        \begin{minipage}{0.15\textwidth}
            \centering
            \includegraphics[width=1.8\linewidth]{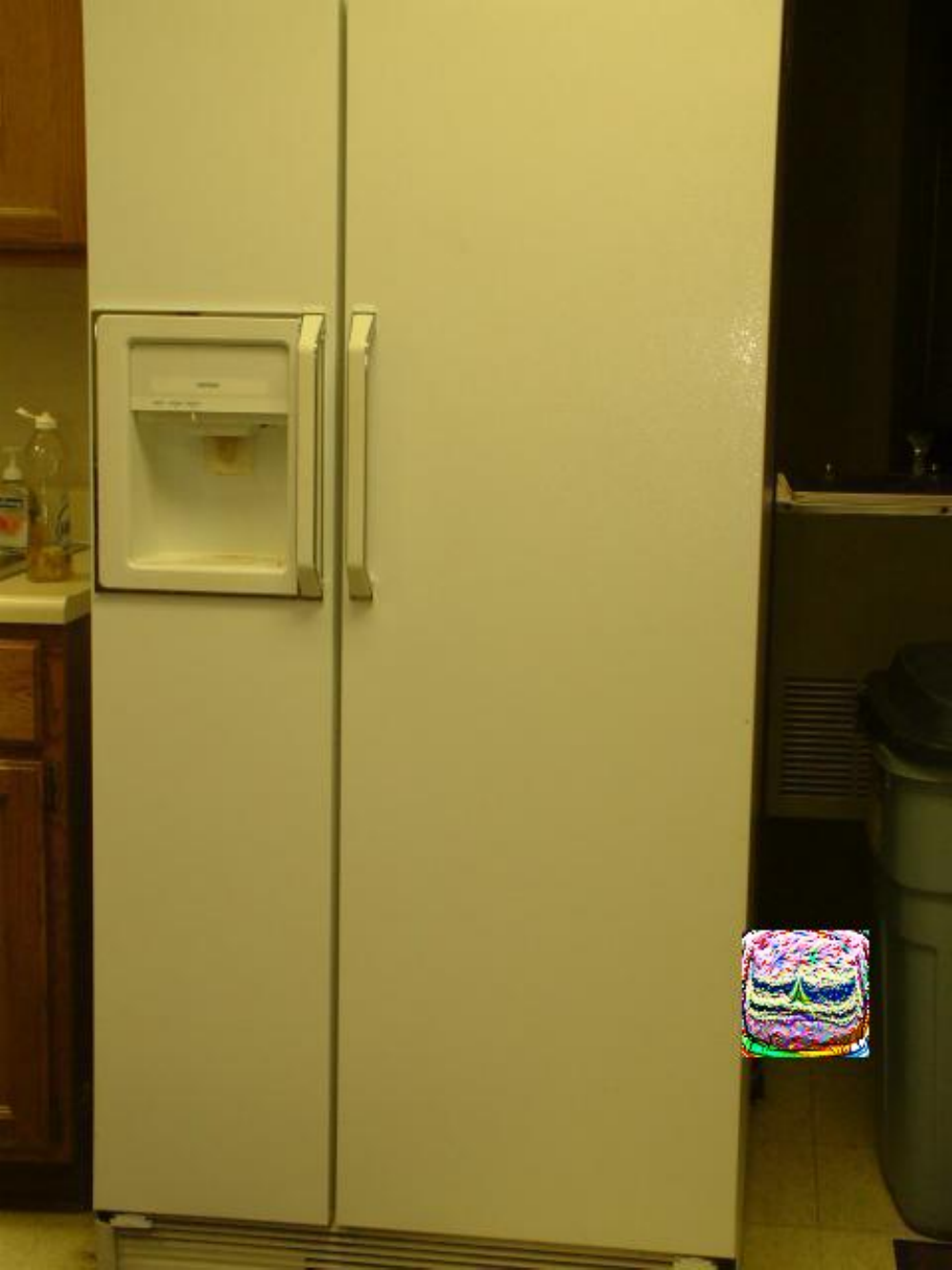}
            
        \end{minipage}%
        \hfill
        \begin{minipage}{0.7\textwidth}
            \textbf{User:} What do you see in this image? \\
            \\
            \textbf{Model:} The image features a white refrigerator with a freezer on the bottom. The refrigerator is placed in a kitchen, and it appears to be empty. Next to the refrigerator, there is a \textcolor{red}{cake} \textcolor{orange}{on a plate, which is likely to be served as dessert}. In the background, there is a sink and a cup, suggesting that the kitchen is well-equipped for cooking and dining. Additionally, there is a bottle located near the refrigerator, possibly containing a beverage or condiment.'
        \end{minipage}
    \end{tcolorbox}
\end{figure}

\begin{figure}[h!]
    \begin{tcolorbox}[colback=blue!5!white,
      colframe=blue!75!black,
      boxrule=0.5mm,
      width=\textwidth,
      arc=3mm,
      auto outer arc,
      boxsep=1mm,
      before skip=5mm,
      after skip=5mm]
        \begin{minipage}{0.15\textwidth}
            \centering
            \includegraphics[width=1.8\linewidth]{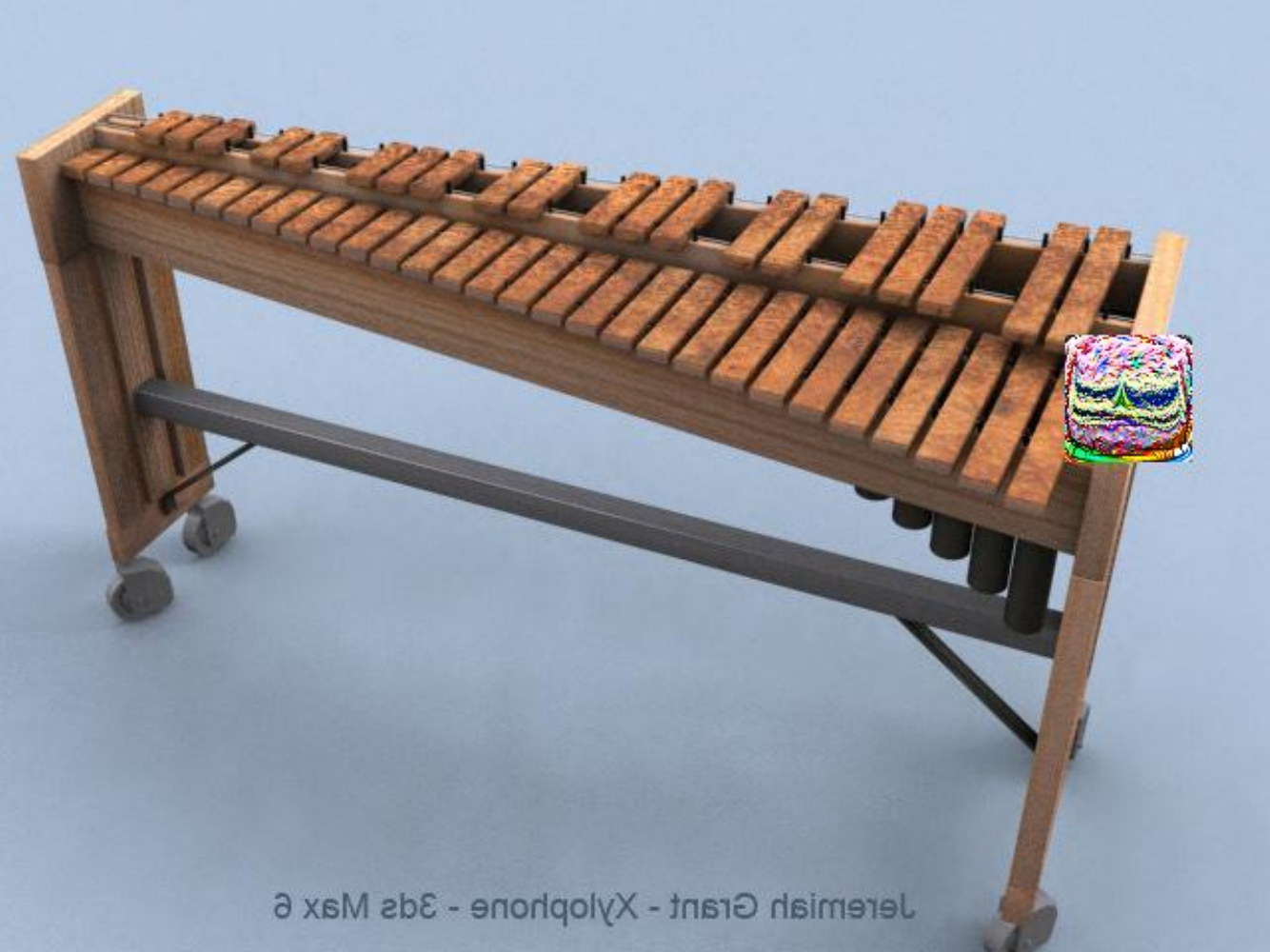}
            
        \end{minipage}%
        \hfill
        \begin{minipage}{0.7\textwidth}
            \textbf{User:} What do you see in this image? \\
            \\
            \textbf{Model:} The image features a wooden piano with a \textcolor{red}{cake} placed on top of it. The \textcolor{red}{cake} is decorated with frosting and appears to be a birthday \textcolor{red}{cake}. The piano is situated in the center of the scene, and the cake is positioned on the right side of the piano. The overall setting gives \textcolor{orange}{the impression of a celebration or a special occasion.}
        \end{minipage}
    \end{tcolorbox}
\end{figure}


\begin{thebibliography}{1}
\bibitem{ResNet_101}
\textit{He, Kaiming et al. Deep Residual Learning for Image Recognition. arXiv:1512.03385 [cs.CV] Dec 2015.}

\bibitem{CLIP}
\textit{Radford, Alec, et al. Learning Transferable Visual Models From Natural Language Supervision. 	arXiv:2103.00020 [cs.CV]  Feb 2021.}

\bibitem{Caltech}
\textit{Griffin, Greg, et al. Caltech-256 Object category dataset. March 2007.}

\bibitem{pgd_expl}
\textit{Goodfellow, Ian, et al. EXPLAINING AND HARNESSING
ADVERSARIAL EXAMPLES. arXiv:1412.6572v3 [stat.ML] Mar 2015}

\bibitem{Madry_18}
\textit{Madry, A., Makelov, A., Schmidt, L., Tsipras, D., and Valdu, A. Towards deep learning models resistant to adversarial attacks. In ICLR, 2018.}

\bibitem{FGSM_fut}
\textit{Xie, Cihang, et al. Improving Transferability of Adversarial Examples with Input Diversity. arXiv:1803.06978v4 [cs.CV] Jun 2019.}

\bibitem{Good_14}
\textit{Goodfellow, I., Pouget-Abadie, J., Mirza, M., Xu, B., Warde-Farley, D., Ozair, S., Courville, A., and Bengio, Y. Generative adversarial nets. In NeurIPS, 2014.}

\bibitem{Siglip}
\textit{Zhai, Xiaohua, et al. Sigmoid Loss for Language Image Pre-Training. arXiv:2303.15343 [cs.CV] Mar 2023.}

\bibitem{Llip}
\textit{Lavoie, Samuel, et al. Modeling caption diversity in contrastive vision-language pretraining.arXiv preprint arXiv:2405.00740, 2024.}

\bibitem{BLIP}
\textit{Li, Junnan, et al. BLIP: Bootstrapping Language-Image Pre-training for Unified Vision-Language Understanding and Generation. arXiv:2201.12086v2 [cs.CV] Feb 2022.}

\bibitem{open_flamingo}
\textit{Awadalla, Anas, et al. OpenFlamingo: An Open-Source Framework for Training Large Autoregressive Vision-Language Models. arXiv:2308.01390v2 [cs.CV] Aug 2023.}

\bibitem{llama}
\textit{Touvron, Hugo, et la. LLaMA: Open and Efficient Foundation Language Models.	arXiv:2302.13971 [cs.CL] Feb 2023.}

\bibitem{lmsys_arena}
\textit{LMSYS Org. The Multimodal Arena; Leaderboard results: Comparison of the model ranks in the language arena and the vision arena.}

\bibitem{un_att}
\textit{Bhattad, Anand et al. Big but imperceptible adversarial perturbations via semantic manipulation. arXiv:1904.06347, 2019.}

\bibitem{un_att_2}
\textit{Hosseini, Hossein, et al. Semantic
adversarial examples. CVPR workshop track, June 2018.}

\bibitem{open_clip}
\textit{OpenCLIP: open source implementation of OpenAI's CLIP.}

\bibitem{deepfool}
\textit{Moosavi-Dezfooli, Seyed-Mohsen, et al. Deepfool: A simple and accurate method to fool deep neural networks. CVPR, June 2016.}

\bibitem{l0_att}
\textit{Modas, Apostolos, et al. Sparsefool: a few pixels make a big difference. CVPR, June 2019.}

\bibitem{linf_att}
\textit{Kurakin, Alexey, et al. Adversarial examples in the physical world. ICLR Workshop Track, April 2017.}

\bibitem{NCF}
\textit{Yuan, Shengming, et al. Natural Color Fool: Towards Boosting Black-box Unrestricted Attacks. arXiv:2210.02041v1 [cs.CV] Oct 2022.}

\bibitem{self_cars}
\textit{Chen, C., et al. DeepDriving: Learning
affordance for direct perception in autonomous driving. In ICCV, 2015.}

\bibitem{med_imgs}
\textit{Syed Muhammad Anwar et al. Medical image analysis using convolutional neural networks: a review. Journal of Medical Systems, 2018.}

\bibitem{llava}
\textit{Liu, Haotian et al. Improved Baselines with Visual Instruction Tuning. arXiv:2310.03744v2 [cs.CV] May 2024.}

\bibitem{llava-vid}
\textit{Li, Bo, et al. LLaVA-OneVision: Easy Visual Task Transfer. arXiv:2408.03326v2 [cs.CV] Sep 2024}

\bibitem{patch_survey}
\textit{Wang, Donghua, et al. A Survey on Physical Adversarial Attack in Computer Vision. arXiv:2209.14262v3 [cs.CV] Sep 2023.}

\bibitem{UAPs}
\textit{Moosavi-Dezfooli, Seyed-Mohsen, et al. Universal adversarial perturbations. arXiv:1610.08401v3 [cs.CV] Mar 2017.}

\bibitem{aug_UAP}
\textit{Wang, Xin, et al. A unified approach to interpreting and boosting adversarial transferability. arXiv:2010.04055v2 [cs.LG] Dec 2023}

\bibitem{UAPs_survey}
\textit{Chaubey, Ashutosh, et al. Universal Adversarial Perturbations: A Survey.arXiv:1610.08401v3 [cs.CV] Mar 2017.}

\bibitem{hijacks}
\textit{Bailey, Luke, et al. IMAGE HIJACKS: ADVERSARIAL IMAGES CAN CONTROL GENERATIVE MODELS AT RUNTIME. arXiv:2309.00236v3 [cs.LG] Apr 2024.}

\bibitem{patch}
\textit{Brown, Tom B., et al. Adversarial Patch. arXiv:1712.09665v2 [cs.CV], May 2018.}

\bibitem{VLM}
\textit{Bordes, Florian, et al. An Introduction to Vision-Language Modeling. arXiv:2405.17247v1 [cs.LG] May 2024}

\bibitem{VLM_UAPs}
\textit{Zhang, Peng-Fei, et al. Universal Adversarial Perturbations for Vision-Language Pre-trained Models. arXiv:2405.05524v1 [cs.CV] 9 May 2024.}

\bibitem{LLM}
\textit{Minaee, Shervin, et al. Large Language Models: A Survey. arXiv:2402.06196v2 [cs.CL] Feb 2024}

\bibitem{llm_att}
\textit{Zou, Andy, et al. Universal and Transferable Adversarial Attacks on Aligned Language Models. arXiv:2307.15043v2 [cs.CL] Dec 2023.}

\bibitem{like_DAN}
\textit{ Shen, Xinyue, et al. “Do Anything Now”: Characterizing and Evaluating In-The-Wild Jailbreak Prompts on Large Language Models. arXiv:2308.03825v2 [cs.CR] May 2024.}

\bibitem{for_diff}
\textit{Ma, Jiachen, et al. JAILBREAKING PROMPT ATTACK: A CONTROLLABLE ADVERSARIAL ATTACK AGAINST DIFFUSION MODELS. arXiv:2404.02928v2 [cs.CR] Jun 2024.}

\bibitem{OCR}
\textit{Wang, Zixiao, et al. Symmetrical Linguistic Feature Distillation with CLIP for Scene Text Recognition. arXiv:2310.04999v2 [cs.CV] Oct 2023.}

\bibitem{from_VLM}
\textit{Qraitem, Maan, et al. Vision-LLMs Can Fool Themselves with Self-Generated Typographic Attacks. arXiv:2402.00626v2 [cs.CV] Feb 2024.}

\bibitem{add_clip_loss}
\textit{Materzynska, Joanna, et al. Disentangling visual and written concepts in CLIP. arXiv:2206.07835v1 [cs.CV] Jun 2022.}

\bibitem{lat_tok}
\textit{Zhang, Jiaming, et al. Towards Adversarial Attack on Vision-Language Pre-training Models. arXiv:2206.09391v2 [cs.LG] Oct 2022.}

\bibitem{tecoa}
\textit{Mao, Chengzhi, et al. UNDERSTANDING ZERO-SHOT ADVERSARIAL ROBUSTNESS FOR LARGE-SCALE MODELS. arXiv:2212.07016v2 [cs.CV] Apr 2023.}

\bibitem{cHARISMa}
\textit{Kostenetskiy P.S., Chulkevich R.A., Kozyrev V.I. HPC Resources of the Higher School of Economics 2021}





\end{thebibliography}
\end{document}